\documentclass[journal]{IEEEtran}
\ifCLASSINFOpdf
\else
\fi
\usepackage{graphicx}
\usepackage{amsmath}
\usepackage{amsfonts}
\usepackage{setspace}
\usepackage{xcolor}

\begin{document}
%
\title{Accelerating the Evolution of Personalized Automated Lane Change through Lesson Learning}
%
%
%

\author{Jia~Hu,~\IEEEmembership{Senior Member,~IEEE}, 
        Mingyue~Lei,
        Haoran~Wang,
        Zeyu Liu,
        Fan Yang
\thanks{Jia Hu, Mingyue Lei and Haoran Wang are with the Key Laboratory of Road and Traffic Engineering of the Ministry of Education, Tongji University, Shanghai 201804, China, e-mail: 
        {\tt\small mingyue\_l@tongji.edu.cn, wang\_haoran@tongji.edu.cn, hujia@tongji.edu.cn}.}
\thanks{Zeyu Liu is with Cooperative Vehicle Infrastructure System (CVIS), Intelligent Transportation Products Department, AI Cloud Group, Baidu Inc., Beijing 100085, China, e-mail: 
        {\tt\small liuzeyu02@baidu.com}.}
\thanks{Fan Yang is with V2X AI Road Open Platform, AI Cloud Group, Baidu Inc., Beijing 100085, China, e-mail: 
        {\tt\small yangfan42@baidu.com}.}}   

\maketitle

\begin{abstract}
Personalization is crucial for the widespread adoption of advanced driver assistance systems. To match up with each user's preference, the online evolution capability is a must. However, conventional evolution methods learn from naturalistic driving data, which requires a lot of computing power and cannot be applied online. To address this challenge, this paper proposes a lesson learning approach: learning from the user's takeover interventions. By learning a lesson from real-time takeover behavior, a driving zone is generated to ensure perceived safety. Within the driving zone, a personalized trajectory is planned based on model predictive control, with an objective learned from the user's takeover. The proposed lesson learning framework is highlighted for its faster evolution capability, adeptness at experience accumulation, assurance of perceived safety, and computational efficiency. Simulation results demonstrate that the proposed system consistently achieves successful customization without further takeover interventions. Accumulated experience yields a 24\% enhancement in evolution efficiency. The average number of learning iterations is only 13.8. The average computation time is 0.08 seconds.
\end{abstract}

\begin{IEEEkeywords}
Human-like driving, Personalization, Model predictive control, Learning based, Automated lane change.
\end{IEEEkeywords}

%
\IEEEpeerreviewmaketitle

\section{Introduction}
%
%
%
%
\subsection{Motivation of Personalized ADAS}
Advanced driver assistance system (ADAS), such as automated lane change (ALC), has emerged as a prominent application in automated driving \cite{wang2023trajectory} \cite{huang2021personalized} \cite{huang2023gameformer} \cite{wang2025field}. The widespread implementation of ADAS still has a long way to go. ADAS has been installed in only 10\% of vehicles globally by the close of 2020 \cite{joe2021only}. Taking Tesla's Full Self-Driving (FSD) as an example, the worldwide order rate for FSD is just 7.4\% as of the third quarter of 2022 \cite{guosen2023special}.

ADAS is now facing the great challenge of frequent takeovers, which reduces its adoption rate \cite{emanuelsson2020examining}. The primary cause of takeovers lies in the lack of personalization. Existing ADAS products are mostly standardized, failing to match up with users' individual preferences, including their personal driving styles and safety expectations \cite{hasenjager2019survey} \cite{yi2019implicit} \cite{li2023safe}. For example, existing ALC systems are typically overly conservative and inefficient compared to an experienced driver \cite{butakov2014personalized}. The mismatch often causes users to manually interrupt the autonomous driving and takeover control authority from ADAS. Hence, personalization is crucial for the promotion of ADAS.

\subsection{Past Studies on Personalized ADAS}
 
Studies have been making efforts on personalized ADAS (PADAS). These studies can be divided into non-learning-based and learning-based approaches.

Non-learning-based approaches primarily consist of statistical-based, optimization-based and numerical-based approaches. In the case of statistical-based approaches, Wang et al. \cite{wang2022gaussian} adopted a Gaussian Process (GP) model to directly learn the correlation between traffic states (gaps and relative speeds) and the ego vehicle's expected future accelerations. Bao et al. \cite{bao2019personalized} adopted Random Forest (RF) methods to assess risk and capture individually preferred travel velocities, enabling personalized safety-focused control for automated vehicles. Wang et al. \cite{wang2018learning} utilized a Gaussian mixture model and hidden Markov model to simulate a driver's individual lane-keeping behavior. Liu et al. \cite{liu2024zone} utilized Markov Decision Processes (MDP) to learn the driver's eye gazing behaviors, capturing the user's visual attention preferences for PADAS. In the case of optimization-based approaches, Yan et al. \cite{yan2022driver} built a merging spot selection model, whose objective function is adjusted by the driver’s aggressiveness factor. Zhang et al. \cite{zhang2022personalized} utilized optimal control to build an obstacle avoidance motion planner. The planner's objective function and constraints follow the needs of passengers. Xu et al. \cite{xu2020learning} established an optimization problem, refining the cost parameters of a longitudinal-lateral coupled motion planner for each individual. In the case of numerical-based approaches, Vigne et al. \cite{vigne2024overtaking} used a parametric sigmoid function to approximate the lane change path, with a single parameter representing the driving style: the lower bound (0) corresponds to a smooth path and a non-aggressive decision-making process, and the upper limit (1) corresponds to a narrower path and an aggressive decision-making process enabling faster overtaking maneuvers. To sum up, these non-learning-based approaches can categorize driving styles into several types. However, their limited number of tunable parameters fails to capture more diverse driving behaviors. Consequently, learning-based approaches have become increasingly critical for PADAS.

Learning-based approaches realize personalization by learning from naturalistic driving data. Various learning methods have been employed for this purpose. Song et al. \cite{song2023personalized} adopted Reinforcement Learning (RL) to model a car-following controller, which adapts to the driver’s desired accelerations and following gap. Zhu et al. \cite{zhu2020combined} used a gated recurrent units (GRU) based combined hierarchy learning framework (CHLF) to plan a personalized lane change trajectory. Huang et al. \cite{huang2021driving} adopted inverse reinforcement learning (IRL) to learn the reward functions of the individual human driver's latent driving intentions. Rosbach et al. \cite{rosbach2019driving} utilized maximum entropy IRL to optimize the driving style of motion planners. Yu et al. \cite{yu2022personalized} utilized batch normalization to build a driver-preference-aware conflict detection classifier, serving for a forward collision warning (FCW) system. Xie et al. \cite{xie2024personalized} adopted a language model to represent the driver's subjective risk level, improving the performance of FCW.
Gao et al. \cite{gao2024adas} adopted spatio-temporal graph transformer networks to predict drivers’ preferences of evasive behavior types, supporting a personalized autonomous evasive takeover (AET) system. Li et al. \cite{li2023research} designed aggressive, calm and moderate braking strategies, utilizing K-means based driving style recognition and encoder-decoder based trajectory prediction. To sum up, thanks to their sufficient tunable parameters, these learning-based approaches have the potential to capture a wider range of driver preferences. However, whether adopting non-learning-based or learning-based approaches, both have a common characteristic: their personalization processes operate in an \textbf{offline} mode. Although some learning-based approaches enable updating after applications (such as operating within a close-loop framework involving application, data collection and evolution), the evolution process remains \textbf{offline}. They require extensive training data and cannot guarantee stability during the training process. Only after collecting enough naturalistic trajectory data can the evolution be achieved.

\subsection{Limitations: Inefficiency of Offline Evolution}
Current ADAS products mostly collect all users' driving data and offline update the product to cover most users' references. However, additional data is essential to extend the range of users' preferences beyond predefined trajectories and enable adaptation to each user's expected driving behaviors. This limitation makes it challenging for current ADAS products to achieve a truly individualized Personalized ADAS (PADAS), which focuses on providing one-on-one personalization tailored to each user. Hence, the offline evolution scheme is not practical. There would be a lot of costs for data transmission and computing, especially when there is a great number of users. To realize the PADAS, online and onboard evolution for each user may be a must.

The key challenge of online evolution is the lack of onboard computing power. The conventional evolution method by nature follows an ``\textbf{imitation learning}" strategy. Personalization is achieved by mimicking human driving behavior based on naturalistic trajectory data. To effectively replicate a human's behavior, extensive driving data must be collected. However, the online training of these massive datasets is impracticable, due to the limitation of onboard computing power. 

\subsection{Proposed Strategy: Lesson Learning to Accelerate Evolution}
To enable the online evolution, we propose a ``\textbf{lesson learning}" strategy. Diverging from learning from massive naturalistic trajectory data in traditional approaches, the proposed ``lesson learning" strategy learns from the user's takeover interventions. Training iteration is activated only when the user takes over the ADAS. The training of massive data is not required anymore. Much less computing power is needed, facilitating online implementation for individual preferences.

\begin{figure}[thpb]
  \centering
  \includegraphics[scale=0.11]{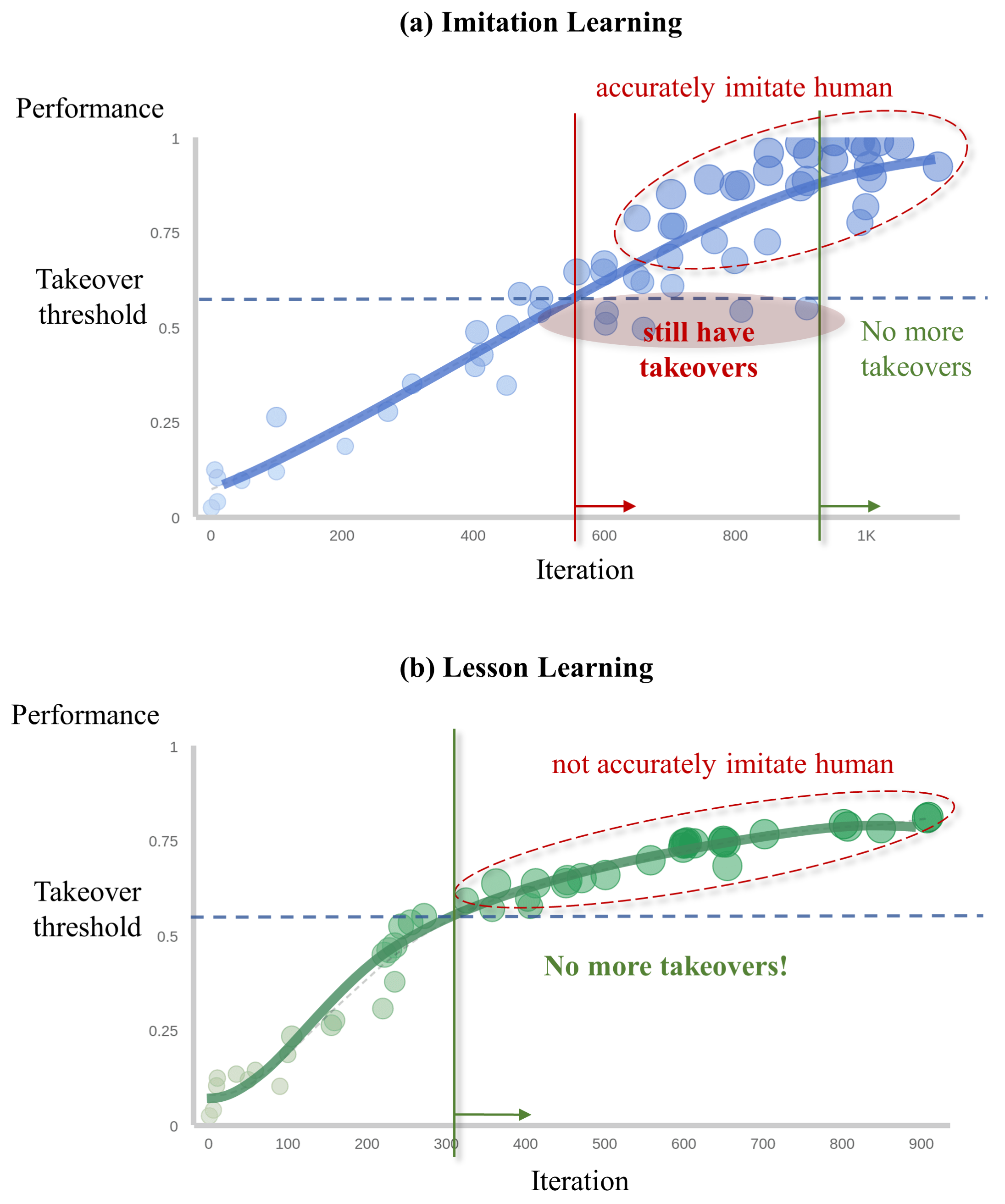}
  \caption{Comparison between ``imitation learning" and ``lesson learning".}
  \label{Figure1}
\end{figure}

Moreover, by its very nature, ``lesson learning" strategy could better align with the objective of PADAS. \textbf{PADAS aims at reducing users' takeover intervention, rather than driving exactly like a human.} The comparison between the results of applying ``imitation learning" and ``lesson learning" is presented in Fig.\ref{Figure1}. The “Performance” axis can be interpreted as the reward for a PADAS operating in a traffic environment. This reward can be measured using various metrics, such as travel efficiency, subjective and objective safety, driving comfort, and so on. The highest reward indicates that the PADAS actions align with human drivers’ desired operations. We assume that actions with rewards above a certain threshold fall within human drivers’ acceptable domain, while those below the threshold are unsatisfactory, requiring human drivers to take over. As iterations progress, “imitation learning” guides the PADAS to perform in a manner that closely matches human drivers’ desired operations (see Fig.\ref{Figure1}(a)). However, due to its stochastic nature, even when the average performance falls within human drivers’ acceptable domain, there may still be outliers that fail to meet human drivers' expectations. Additional iterations are needed to ensure that the PADAS actions fully converge around human drivers’ ideal operations. On the other hand, “lesson learning” guides the PADAS to perform within human drivers' acceptable domain, rigorously constraining outliers (see Fig.\ref{Figure1}(b)). This approach leads to faster convergence toward satisfying human drivers, even though the PADAS may not perform exactly as human drivers would ideally desire. In summary, the ``lesson learning" strategy is capable of reducing iterations and accelerating the evolution.

In this paper, we propose a "lesson learning" based PADAS controller, taking the personalized automated lane change (PALC) scenario as a case study. It bears the following contributions:

\begin{itemize}

\item \textbf{With faster evolution capability}: The proposed controller, utilizing the "lesson learning" strategy, evolves through the learning of human takeover interventions. \textit{Instead of aiming at best matching the user's expectation, the proposed controller is with the objective of finding the user's acceptable domain}. Hence, it needs less training iterations and training data.
\item \textbf{With experience accumulation capability for expanded applicability}: The user's preference is modeled as the preferred driving zone in a relative space. \textit{The proposed system updates the control policy from determining "state-action" pairs to "state-driving zone" pairs}. The formulation of the driving zone is the experience that could be adopted in a new driving environment and accelerating evolutions.
\item \textbf{With perceived safety ensured}: The proposed PALC system is engineered to minimize user's interventions. To achieve this goal, the system adopts the avoidance of perceived unsafe zones, where takeovers frequently occur, as state constraints for the planner. Over a series of iterations, a \textit{perceived safe driving zone} is established, contributing to a heightened sense of safety for the system's operation.
\item \textbf{With the capability of real-time field implementation}: The proposed system facilitates real-time onboard computing by learning from the takeover behavior, instead of massive naturalistic driving data. Furthermore, the "lesson learning" strategy enables faster evolution with \textit{less training iterations}. Consequently, the onboard computation power is sufficient to support the system's operation.

\end{itemize}

\section{METHODOLOGY}
The goal of this paper is to develop a PALC system, that enables online personalization to individual user's expectation. The evolution is achieved by iteratively updating the controller by capturing the discrepancy between the user's acceptable domain and the ego vehicle's motions. The user's acceptable domain is inferred from the user's takeover interventions.

\subsection{Lesson Learning Logic}
\begin{figure}[thpb]
  \centering
  \includegraphics[scale=0.45]{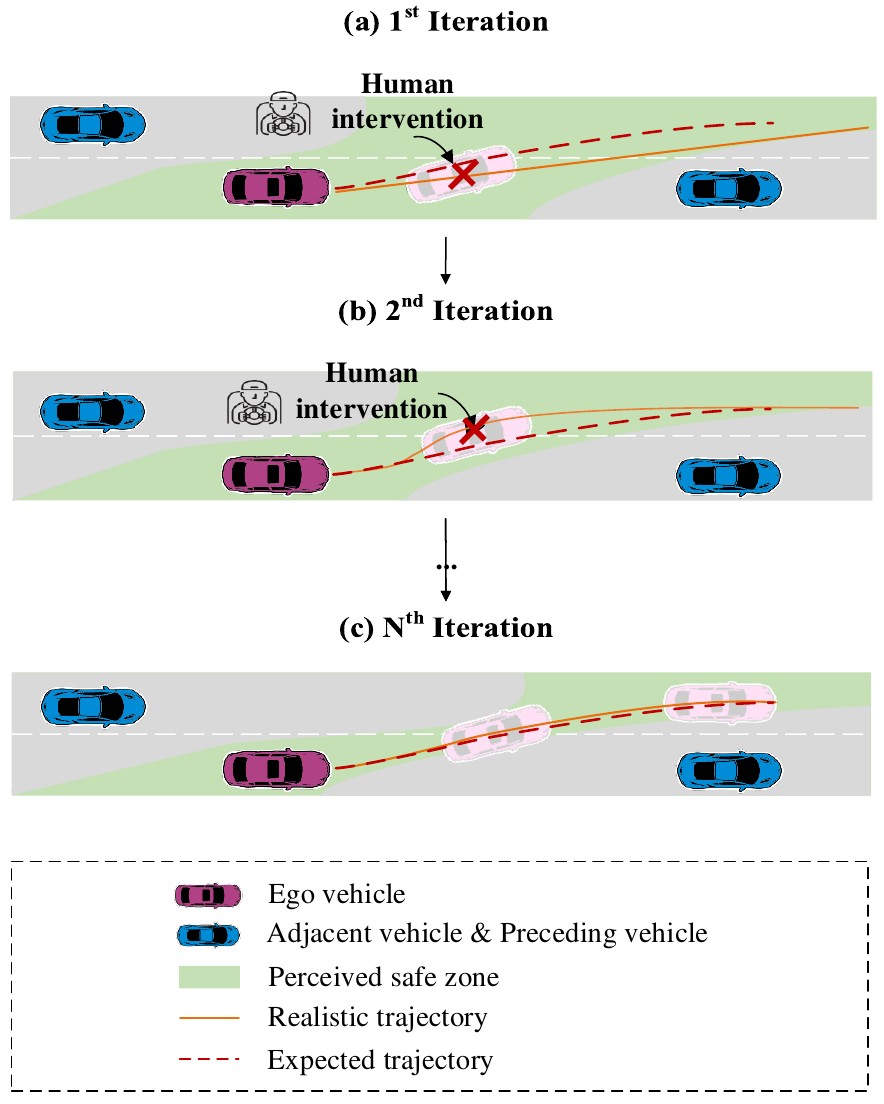}
  \caption{The logic of lesson learning.}
  \label{Figure2}
\end{figure}

Lesson learning strategy operates as an iteration process while driving, as shown in Fig.\ref{Figure2}. Each iteration is triggered by human intervention and operated online. Within each iteration, the perceived safe zone and the expected trajectory are updated based on the user's intervention behavior, as shown in Fig.\ref{Figure2}(a)-Fig.\ref{Figure2}(b). The perceived safe zone is trained to match up with human's expectations. The expected trajectory is planned within the perceived safe zone. Through repeated iterations, the trained perceived safe zone shrinks to the user's expectations, as shown in Fig.\ref{Figure2}(c), ultimately avoiding more interventions and culminating in the achievement of customization.

\subsection{System Architecture}
\begin{figure*}[!t]
  \centering
  \includegraphics[width=\linewidth]{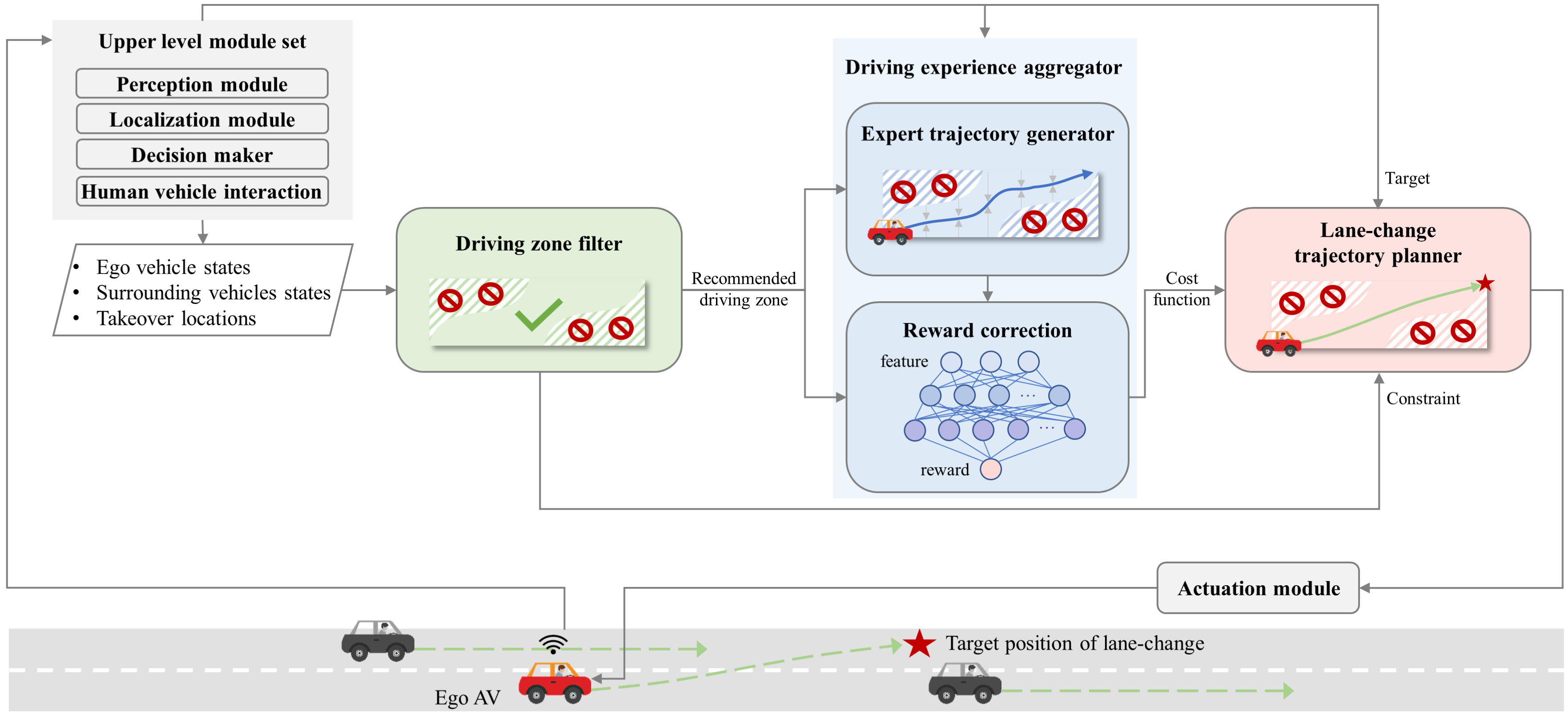}
  \caption{Framework of the proposed system.}
  \label{Figure3}
\end{figure*}

To realize the lesson learning strategy, the framework of the proposed PADAS system is designed as shown in Fig.\ref{Figure3}. It consists of five modules. The focus of this research is on the development of the driving zone filter, the driving experience aggregator and the lane-change trajectory planner. The details of the framework are provided as follows:

\subsubsection{Upper level module set}
The upper-level module serves as the input source for the proposed motion planner, comprising three primary sub-modules: perception, localization, and decision-making. The perception module collects information regarding surrounding traffic conditions. The localization module determines the state of the ego vehicle. Lastly, the decision-maker module provides the target position and target velocity required for lane-change maneuvers.

\subsubsection{Driving zone filter}
This module generates a recommended driving zone, which serves as a boundary constraint for the planning of the ego vehicle. The recommended driving zone is iteratively adjusted according to the human's takeover interventions. By confining trajectory planning within this driving zone, the system ensures perceived safety and reduces the frequency of takeovers.

\subsubsection{Driving experience aggregator}
This module aims to tailor the driving experience to match the user's individual style by aggregating their driving behavior from two aspects: expert trajectory generation and planning reward correction. Expert trajectories are generated based on the user's preference derived from intervention data and the recommended driving zone. The planning reward correction updates the reward function used in planning to mimic the behavior of the expert, thereby ensuring alignment with the user's driving style and preferences.

\subsubsection{Lane-change trajectory planner}
This module plans personalized lane-change trajectory, utilizing the trained reward function provided in the driving experience aggregator module. Perceived safety is ensured via planning within the recommended driving zone. 

\subsubsection{Actuation module}
This module executes commands generated by the planner through local control mechanisms. In the event of a user takeover, a new iteration for system upgrading is triggered.

\subsection{Problem Formulation}
Based on the aforementioned logic and system architecture, the motion planning problem is formulated for PADAS. ALC trajectory is computed by an optimal control problem. The control objective is to follow an expert's driving style, which is derived by aggregating the user's driving experience. The planning problem is constrained by a perceived safe driving zone, which is derived by a driving zone filter.

\subsubsection{Takeover Acquisition for Filtering Driving Zone}

The training data of the PADAS system is the positions and the takeover locations of vehicles including both the ego vehicle and its surrounding vehicles. The takeover location refers to the position of the ego vehicle when the human driver takes over control from the PADAS.

The collected takeover data are used for shaping a driving zone for the ego vehicle. A driving zone filter is presented to generate the driving zone for automated vehicles. Driving within this zone, the automated vehicles have a low possibility of being overrode by user.

\begin{figure}[thpb]
  \centering
  \includegraphics[scale=0.12]{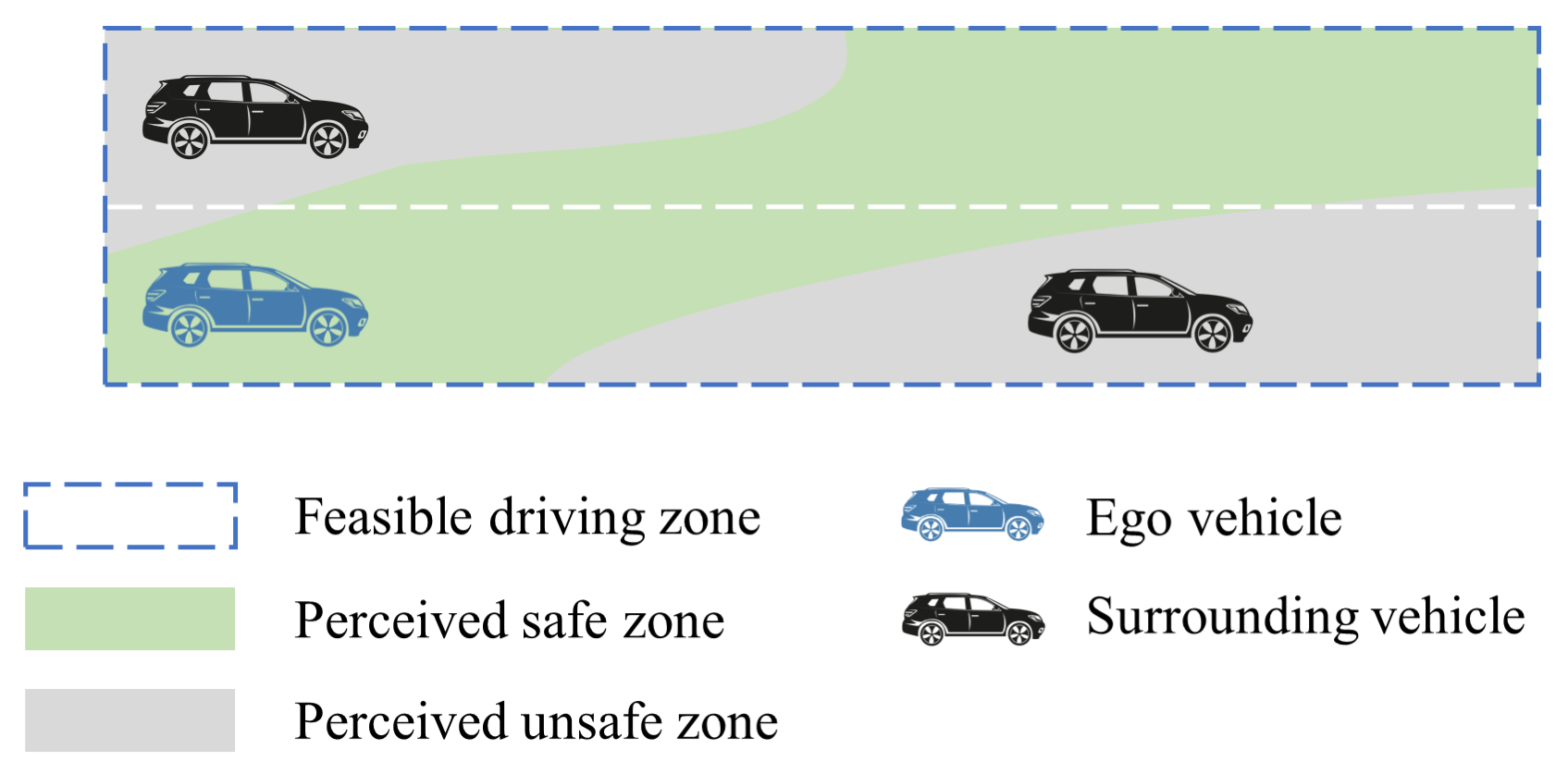}
  \caption{Three types of driving zones.}
  \label{Figure4}
\end{figure}

Three types of driving zones are defined, as shown in Fig.\ref{Figure4}:

\begin{itemize}

\item \textbf{Feasible driving zone}: The zone where the ego vehicle can reach in the future control horizon. It is divided into perceived safe zone and perceived unsafe zone.
\item \textbf{Perceived safe zone}: The zone where the ego vehicle has a higher belief probability of not being taken over by the human driver than that of being taken over.
\item \textbf{Perceived unsafe zone}: The zone where the ego vehicle has a higher belief probability of being taken over by the user than that of not being taken over.

\end{itemize}

The perceived safe zone is the recommended driving zone presented in the preceding sections. To obtain the proposed perceived safe zone, a gaussian discriminant analysis (GDA) based position classification method is designed as follows.

Feature set $\boldsymbol{x}$ is defined to represent the relative state in the background traffic. 
\begin{equation}
    \boldsymbol{x}=\left[\begin{array}{llllll}
    \Delta s_{p} & \Delta l_{p} & \operatorname{dist}^{p} & \Delta s_{a} & \Delta l_{a} & \text {dist}^{a}
    \end{array}\right]^{T}
    \label{Equation1}
\end{equation}
where $\Delta s_{p}$ is the relative longitudinal position between the ego vehicle and its proceeding one, $\Delta l_{p}$ is the relative lateral position between the ego vehicle and its proceeding one, $\operatorname{dist}^{p}$ is the distance between the ego vehicle and its proceeding one, $\Delta s_{a}$ is the relative longitudinal position between the ego vehicle and the adjacent vehicle on the target lane, $\Delta l_{a}$ is the relative lateral position between the ego vehicle and the adjacent vehicle on the target lane, $\text {dist}^{a}$ is the distance between the ego vehicle and the adjacent vehicle on the target lane.

The class of a position with feature $\boldsymbol{x}$ is defined as $y$. Specifically, $y=0$ means that the ego vehicle is taken over by the user, and $y=1$ means that the ego vehicle is not taken over by the user.

The classification model is formulated based on the Bayes rule \cite{efron2013bayes} as follows.
\begin{equation}
    \begin{array}{c}
    \operatorname{class}(\boldsymbol{x})=\underset{{y}}{\operatorname{argmax}} p(y\! \mid \!\boldsymbol{x})=\underset{{y}}{\operatorname{argmax}} \frac{p(\boldsymbol{x} \mid y) p(y)}{p(\boldsymbol{x})} \\
    =\underset{y}{\operatorname{argmax}} p(\boldsymbol{x}\! \mid \!y) p(y)
    \end{array}
    \label{Equation2}
\end{equation}
where $\underset{y}{\operatorname{argmax}} p(y\! \mid \!\boldsymbol{x})=1$ means that the ego vehicle has a higher belief probability of not being taken over by the user than that of being taken over given $\boldsymbol{x}$. Hence, $class(\boldsymbol{x})=1$ means that the ego vehicle located at the given situation $\boldsymbol{x}$ is in the perceived safe zone.

$p(\boldsymbol{x}\! \mid \!y)$ is the condition distribution of $y$ given $\boldsymbol{x}$. It is assumed to be a multivariate Gaussian distribution as follows.
\begin{equation}
    p(\boldsymbol{x} ; \boldsymbol{\mu}, \boldsymbol{\Sigma} \mid y)=\frac{1}{\sqrt{(2 \pi)^{n}|\boldsymbol{\Sigma}|}} e^{-\frac{1}{2}(\boldsymbol{x}-\boldsymbol{\mu})^{T} \boldsymbol{\Sigma}^{-1}(\boldsymbol{x}-\boldsymbol{\mu})}
    \label{Equation3}
\end{equation}
where $\boldsymbol{\mu}$ is the mean vector, $\mathbf{\Sigma}$ is the covariance matrix, and $n$ is the dimension of $\boldsymbol{x}$. Specifically, in the case of $y=1$, the corresponding mean vector is defined as $\boldsymbol{\mu}_{1}$. In the case of $y=0$, the corresponding mean vector is defined as $\boldsymbol{\mu}_{0}$.

$p(y)$ is class prior distribution. As $y$ either takes the value 1 or 0, it is assumed to be a Bernoulli distribution as follows.

\begin{equation}
    p(y ; \theta)=\theta^{y}(1-\theta)^{1-y}, y \in\{0,1\}
    \label{Equation4}
\end{equation}
where $\theta$ is the possibility of $y=1$.

The values of the parameters $\theta$, $\boldsymbol{\mu}$ and $\mathbf{\Sigma}$ are estimated via Maximum Likelihood estimation \cite{myung2003tutorial}. The data utilized for estimation are collected from historical driving trajectories that ultimately result in takeovers. In the case that the ego vehicle is operated by automated driving system, the corresponding $y$ equals to 1. In the case that the ego vehicle is operated by user, the corresponding $y$ equals to 0.
\begin{equation}
    \theta=\frac{N_{1}}{N}
    \label{Equation5}
\end{equation}
\begin{equation}
    \boldsymbol{\mu}_{1}=\frac{\sum_{i=1}^{N} y_{i} \boldsymbol{x}_{i}}{N_{1}}
    \label{Equation6}
\end{equation}
\begin{equation}
    \boldsymbol{\mu}_{0}=\frac{\sum_{i=1}^{N} y_{i} \boldsymbol{x}_{i}}{N_{0}}
    \label{Equation7}
\end{equation}
\begin{equation}
    \begin{array}{l}
    \boldsymbol{\Sigma}=
    \frac{N_{1} \sum_{i=1}^{N}\left(\boldsymbol{x}_{i}-\boldsymbol{\mu}_{1}\right)^{T}\left(\boldsymbol{x}_{i}-\boldsymbol{\mu}_{1}\right)+N_{0} \sum_{i=1}^{N}\left(\boldsymbol{x}_{i}-\boldsymbol{\mu}_{0}\right)^{T}\left(\boldsymbol{x}_{i}-\boldsymbol{\mu}_{0}\right)}{N^{2}}
    \end{array}
    \label{Equation8}
\end{equation}
where $(\boldsymbol{x}_i,y_i)$ is the $i^{t h}$ data sample, $N$ is the total number of data sample, $N_1$ is the number of data sample with $y=1$, $N_0$ is the number of data sample with $y=0$. 

\subsubsection{Trajectory Planner}

The system state of PADAS includes the ego vehicle’s positions, speeds and heading angles throughout the future control horizon. A trajectory planner is developed to optimize these states.

\paragraph{State explanation}

The system state vector $\boldsymbol{X}_{i}$ is defined as follows.
\begin{equation}
    \boldsymbol{X}_{i}=\left[\begin{array}{llll}
    s_{i} & v_{i} & l_{i} & \varphi_{i}
    \end{array}\right]^{T}, i \in[0, K]
    \label{Equation9}
\end{equation}
where $s_i$ is the ego vehicle's longitudinal position at step $i$, $v_i$ is the ego vehicle's desired longitudinal speed at step $i$, $l_i$ is the ego vehicle's lateral position at step $i$, $\varphi_i$ is the ego vehicle's heading angle at step $i$, and $K$ is the control horizon. During implementation, $K$ is approximated according to the takeover positions of the ego vehicle. The objective is to provide enough steps for compensating the lateral distance between (1) the center line at the narrowest section of the perceived safe zone and (2) the ego vehicle's target lateral position at the end of the lane-change maneuver.

The control vector $\boldsymbol{U}_{i}$ is defined as follows.
\begin{equation}
    \boldsymbol{U}_{i}=\left[\begin{array}{ll}
    a_{i} & \delta_{i}
    \end{array}\right]^{T}, i \in[0, K-1]
    \label{Equation10}
\end{equation}
where $a_i$ is the ego vehicle's acceleration at step $i$, $\delta_{i}$ is the ego vehicle's front wheel angle at step $i$.

\paragraph{Vehicle dynamics}

Vehicle dynamics model is formulated as follows.
\begin{equation}
    \boldsymbol{X}_{i+1}=\boldsymbol{A}_{i} \boldsymbol{X}_{i}+\boldsymbol{B}_{i} \boldsymbol{U}_{i}+\boldsymbol{C}_{i}, i \in[0, K-1]
    \label{Equation11}
\end{equation}
with
\begin{equation}
    \boldsymbol{A}_{i}=\Delta t \times\left[\begin{array}{cccc}
    0 & 0 & 1 & 0 \\
    0 & 0 & 0 & v_{i} \\
    0 & 0 & 0 & 0 \\
    0 & 0 & 0 & 0
    \end{array}\right]+\boldsymbol{I}_{4 \times 4}, i \in[0, K-1]
    \label{Equation12}
\end{equation}
\begin{equation}
    \boldsymbol{B}_{i}=\Delta t \times\left[\begin{array}{cc}
    0 & 0 \\
    0 & 0 \\
    1 & 0 \\
    0 & \frac{v_{i}}{l_{f r}}
    \end{array}\right], i \in[0, K-1]
    \label{Equation13}
\end{equation}
\begin{equation}
    \boldsymbol{C}_{i}=\Delta t \times\left[\begin{array}{cc}
    0 & 0 \\
    0 & 0 \\
    0 & 0 \\
    0 & -v_{i} R
    \end{array}\right], i \in[0, K-1]
    \label{Equation14}
\end{equation}
where $\Delta t$ is the time increment in each step, $l_{f r}$ is the distance between the ego vehicle's front axle and rear axle, and $R$ is the road curvature.

\paragraph{Cost function}

The planning objective is to follow an expert trajectory, which is formulated into a reward function as follows. This reward function could be updated via aggregating the user's driving experience (see Section \uppercase\expandafter{\romannumeral2}.C.3).
\begin{equation}
    \begin{array}{c}
    J=\sum_{j=1}^{10} \sum_{i=1}^{K} w_{j, i} \cdot \boldsymbol{\Phi}_{j, i}(\boldsymbol{X}) \\
    =\sum_{i=1}^{K}\left(w_{1, i} l_{i-1} l_{i-1}+w_{2, i} l_{i-1}+w_{3, i} \varphi_{i-1} \varphi_{i-1}\right. \\
    +w_{4, i} \varphi_{i-1}+w_{5, i} l_{i-1} \delta_{i-1}+w_{6, i} \varphi_{i-1} \delta_{i-1}+w_{7, i} s_{i-1} \delta_{i-1} \\
    \left.+w_{8, i} \delta_{i-1} \delta_{i-1}+w_{9, i} d i s t_{i-1}^{p}+w_{10, i} d i s t_{i-1}^{a}\right) \\
    =\sum_{i=1}^{K}\left(\boldsymbol{X}_{i-1}^{T} \boldsymbol{Q}_{i} \boldsymbol{X}_{i-1}+\boldsymbol{D}_{i} \boldsymbol{X}_{i-1}\right. \\
    +\boldsymbol{X}_{i-1}^{T} \boldsymbol{F}_{i} \boldsymbol{U}_{i-1}+\boldsymbol{U}_{i-1}^{T} \boldsymbol{R}_{i} \boldsymbol{U}_{i-1})
    \end{array}
    \label{Equation15}
\end{equation}
with
\begin{equation}
    \begin{array}{c}
    \boldsymbol{Q}_{i}=\left[\begin{array}{cccc}
    w_{9, i}+w_{10, i} & 0 & 0 & 0 \\
    0 & 0 & 0 & 0 \\
    0 & 0 & w_{1, i}+w_{9, i}+w_{10, i} & 0 \\
    0 & 0 & 0 & w_{3, i}
    \end{array}\right], \\
    i \in[1, K]
    \end{array}
    \label{Equation16}
\end{equation}
\begin{equation}
    \begin{array}{l}
    \begin{array}{l}
    \boldsymbol{D}_{i}=\left[\begin{array}{c}
    -2 w_{9, i} s_{i-1}^{p}-2 w_{10, i} s_{i-1}^{a} \\
    0 \\
    w_{2, i}-2 w_{9, i} l_{i-1}^{p} \\
    -2 w_{10, i} l_{i-1}^{a} w_{4, i}
    \end{array}\right], i \in[1, K] \\
    \end{array}
    \end{array}
    \label{Equation17}
\end{equation}
\begin{equation}
    \boldsymbol{F}_{i}=\left[\begin{array}{cccc}
    0 & 0 & 0 & 0 \\
    w_{7, i} & 0 & w_{5, i} & w_{6, i}
    \end{array}\right]^{T}, i \in[1, K]
    \label{Equation18}
\end{equation}
\begin{equation}
    \boldsymbol{R}_{i}=\left[\begin{array}{cc}
    0 & 0 \\
    0 & w_{8, i}
    \end{array}\right], i \in[1, K]
    \label{Equation19}
\end{equation}
where $\boldsymbol{w} \in R^{10 \times K}$ is the weighting matrix produced by the proposed reward correction module (see Section \uppercase\expandafter{\romannumeral2}.C.3), $\boldsymbol{\Phi}(\boldsymbol{X}) \in R^{10 \times K}$ is the feature expectation of $\boldsymbol{X}$, $w_{j,i}$ is the $(j,i)^{t h}$ element of $\boldsymbol{w}$, $\operatorname{dist}_{i}^{p}$ is the distance between the ego vehicle and its preceding one at step $i$, $\operatorname{dist}_{i}^{a}$ is the distance between the ego vehicle and the adjacent vehicle on the target lane at step $i$, $s_i^p$ is the longitudinal position of the ego vehicle's preceding vehicle at step $i$, $s_i^a$ is the longitudinal position of the ego vehicle's adjacent vehicle on the target lane at step $i$, $l_i^p$ is the lateral position of the ego vehicle's preceding vehicle at step $i$, and $l_i^a$ is the lateral position of the ego vehicle's adjacent vehicle on the target lane at step $i$.

The motivation for choosing this specific formulation of the cost function is that it enables accurate trajectory planning compared to low-parametrized models. For example, if a spline-based model is adopted, only the parameters of the position and speed at the lane change endpoint can be adjusted. It indicates that all lane change trajectories with identical endpoint states would necessarily be the same. However, in real-world scenarios, human drivers' lane change maneuvers are influenced by various additional factors, such as the positions where they cross lane markings. It results in more diverse trajectory patterns. This motivates the inclusion of additional learnable parameters in the cost function to better capture the full range of possible lane change trajectories.

\paragraph{Boundary conditions}

The initial state is known and acquired via localization:
\begin{equation}
    \boldsymbol{X}_{0}=\left[\begin{array}{llll}
    s_{0} & v_{0} & l_{0} & \varphi_{0}
    \end{array}\right]^{T}
    \label{Equation20}
\end{equation}
where $s_0$ is the ego vehicle's current longitudinal position, and $v_0$ is the ego vehicle's current speed, $l_0$ is the ego vehicle's current lateral position, and $\varphi_{0}$ is the ego vehicle's current heading angle.

The terminal state is restrained, which is determined by the upper decision maker:
\begin{equation}
    \boldsymbol{X}_{K}=\left[\begin{array}{llll}
    s_{K}^{d} & v_{K}^{d} & l_{K}^{d} & 0
    \end{array}\right]^{T}
    \label{Equation21}
\end{equation}
where $s_K^d$ is the target longitudinal position at the end of the lane-change maneuver, $v_K^d$ is the target speed after the lane-change, and $l_K^d$ is the target lateral position at the end of the lane-change maneuver.

\paragraph{Constraints}

Ego vehicle is not allowed to travel outside the proposed recommended driving zone. The recommended driving zone has been presented in Section \uppercase\expandafter{\romannumeral2}.C.1.
\begin{equation}
    l_{i, \min } \leq l_{i} \leq l_{i, \max }, i \in[0, K]
    \label{Equation22}
\end{equation}
where $l_{i,min}$ is the minimum allowed lateral position at step $i$, and $l_{i,max}$ is the maximum lateral position at step i. $l_{i,min}$ and $l_{i,max}$ are obtained according to the perceived safe zone in the Section \uppercase\expandafter{\romannumeral2}.C.1, with detailed calculations provided in equation (\ref{Equation27}) - (\ref{Equation32}).

Ego vehicle's front wheel angle should be limited considering vehicle's capability:
\begin{equation}
    \delta_{\min } \leq \delta_{i} \leq \delta_{\max }, i \in[0, K-1]
    \label{Equation23}
\end{equation}
where $\delta_{\min }$ is the minimum front wheel angle, and $\delta_{\max }$ is the maximum front wheel angle.

\paragraph{Solution method}

A Pontryagin’s Minimum Principle (PMP) based method is utilized to obtain the optimal control law of the proposed lane-change trajectory planner. The detailed solution is previously developed by this research team \cite{wang2022make} \cite{hu2023lane}.

\subsubsection{Driving Experience Aggregator}

The learning algorithm utilized in PADAS system is apprenticeship learning \cite{abbeel2004apprenticeship}. It is a learning-from-demonstration method, with the objective of reconstructing the control policy. The expert demonstrations in apprenticeship learning are not ground truth labels, but approximations derived from the training data. The system states in the Section \uppercase\expandafter{\romannumeral2}.C.2 are updated by the learning algorithm in the Section \uppercase\expandafter{\romannumeral2}.C.3. The system interacts with environment, and generates additional training data for subsequent learning iterations. To realize the learning process, a driving experience aggregator is designed to generate the expert trajectory and update the reward function in Section \uppercase\expandafter{\romannumeral2}.C.2.

\paragraph{Expert trajectory generator}

The user's experience is aggregated into an expert trajectory following the lesson learning strategy. It is unrealistic to directly obtain the user's desired driving trajectory. Hence, in the proposed system, the expert trajectory is inferred from the user's reaction to unexpected vehicle motions.

A model predictive control based method is utilized to realize the process of expert trajectory generation. The expert trajectory generating objective is to follow the user's preference (represented by $C^{prefer}$), at the same time rewarding existing expectations (represented by $C^{init}$). The cost function of expert $C^E$ is formulated as follows.
\begin{equation}
    C^{E}=C^{\text {init }}+C^{\text {prefer }}
    \label{Equation24}
\end{equation}
where $C^{init}$ is the cost function that has been adopted in the proposed trajectory planner, and $C^{prefer}$ is the cost of following the user's preference. The form of $C^{init}$ is the same as the form of equation (\ref{Equation15}).

$C^{prefer}$ is modeled by penalizing the motions that are away from the user's recommended driving zone. The recommended driving zone has been presented in Section \uppercase\expandafter{\romannumeral2}.C.1. The magnitude of the penalty is quantified by the distance between the ego vehicle's trajectory and the center line of the user's recommended driving zone:
\begin{equation}
    C^{\text {prefer }}=\sum_{i=1}^{K}\left(\boldsymbol{X}_{i}-\boldsymbol{X}_{i}^{p}\right)^{T} \boldsymbol{Q}^{p}\left(\boldsymbol{X}_{i}-\boldsymbol{X}_{i}^{p}\right)
    \label{Equation25}
\end{equation}
\begin{equation}
    \boldsymbol{X}_{i}^{p}=\left[\begin{array}{llll}
s_{i} & v_{i} & \frac{\left(l_{i, \min }+l_{i, \max }\right)}{2} & \varphi_{i}
\end{array}\right]^{T}, i \in[1, K]
    \label{Equation26}
\end{equation}
where $\boldsymbol{X}_{i}$ has the same explanation as that in equation (\ref{Equation9}), $\boldsymbol{X}_i^p$ is the ego vehicle's target state at step $i$, $s_i$, $v_i$ and $\varphi_{i}$ in $\boldsymbol{X}_i^p$ have the same explanations as those variables in equation (\ref{Equation9}), $\boldsymbol{Q}^{p}$ is the weighting matrix, $l_{i,min}$ and $l_{i,max}$ have the same explanations as those variables in equation (\ref{Equation22}). Tracking $\boldsymbol{X}^{p}$ contributes to reducing the difference between $y_i$ and $(l_{i,min}+l_{i,max} )/2$, which guiding the ego vehicle to be kept within the scope of the user's preference.

$l_{i,min}$ and $l_{i,max}$ are obtained by projecting the recommended driving zone's boundary onto the coordinate of the vehicular control system.
\begin{equation}
    \Pi_{i}^{-}=\left\{j \mid \boldsymbol{B}_{2, j}<l_{i}, \forall j \in[1, J]\right\}, i \in[1, K]
    \label{Equation27}
\end{equation}
\begin{equation}
    \Pi_{i}^{+}=\left\{j \mid \boldsymbol{B}_{2, j}>l_{i}, \forall j \in[1, J]\right\}, i \in[1, K]
    \label{Equation28}
\end{equation}
\begin{equation}
    j j_{i}^{-}=\arg \min _{j \in \Pi_{i}^{-}}\left(\left|\boldsymbol{B}_{1, j}-s_{i}\right|\right), i \in[1, K]
    \label{Equation29}
\end{equation}
\begin{equation}
    j j_{i}^{+}=\arg \min _{j \in \Pi_{i}^{+}}\left(\left|\boldsymbol{B}_{1, j}-s_{i}\right|\right), i \in[1, K]
    \label{Equation30}
\end{equation}
\begin{equation}
    l_{i, \min }=\boldsymbol{B}_{2, j j_{i}^{-}}, i \in[1, K]
    \label{Equation31}
\end{equation}
\begin{equation}
    l_{i, \max }=\boldsymbol{B}_{2, j j_{i}^{+}}, i \in[1, K]
    \label{Equation32}
\end{equation}
where $\boldsymbol{B} \in R^{2 \times J}$ is a set of points that form the boundary of the recommended driving zone, $J$ is the number of points in $\boldsymbol{B}$. $\boldsymbol{B}_{1,j}$ is the longitudinal coordinate of the $j^{t h}$ point in $\boldsymbol{B}$, and $\boldsymbol{B}_{2,j}$ is the lateral coordinate of the $j^{t h}$ point in $\boldsymbol{B}$. $\Pi_{i}^{-}$ is the set that contains all the indexes of the points that with smaller lateral coordinate than that of the ego vehicle at step $i$, and $\Pi_{i}^{+}$ is the set that contains all the indexes of the points that with greater lateral coordinate than that of the ego vehicle at step $i$. $j j_{i}^{-}$ is the index of the point that is closest to the ego vehicle among $\Pi_{i}^{-}$, and $j j_{i}^{+}$ is the index of the point that is closest to the ego vehicle among $\Pi_{i}^{+}$.

\paragraph{Reward correction}

Reward of the trajectory planner is updated to adapt to the user's preference. The correction function is formulated based on apprenticeship learning \cite{abbeel2004apprenticeship}.
\begin{equation}
    \boldsymbol{w}=\arg \min _{\boldsymbol{w}} \max _{\boldsymbol{\psi}}\left(\boldsymbol{w} \cdot \boldsymbol{\Phi}(\boldsymbol{\psi})-\boldsymbol{w} \cdot \boldsymbol{\Phi}\left(\boldsymbol{\psi}^{\boldsymbol{E}}\right)\right)
    \label{Equation33}
\end{equation}
where $\boldsymbol{w} \in R^{M \times K}$ is the weighting matrix, giving a weight to each feature. $M$ is the number of features captured at one single step in a trajectory, and $K$ is the number of steps in a trajectory. $\boldsymbol{\psi}$ is the planned trajectory. $\Phi(\boldsymbol{\boldsymbol{\psi}}) \in R^{M \times K}$ is the feature expectation of $\boldsymbol{\psi}$. $\boldsymbol{\psi}^E$ is the expert trajectory, which is provided by the proposed expert trajectory generator. $\min _{\boldsymbol{w}}\left(\boldsymbol{w} \cdot \boldsymbol{\Phi}(\boldsymbol{\psi})-\boldsymbol{w} \cdot \boldsymbol{\Phi}\left(\boldsymbol{\psi}^{E}\right)\right)$ contributes to the cost enhancement when making deviations from expert. $\max _{\boldsymbol{\psi}}\left(\boldsymbol{w} \cdot \boldsymbol{\Phi}(\boldsymbol{\psi})-\boldsymbol{w} \cdot \boldsymbol{\Phi}\left(\boldsymbol{\psi}^{E}\right)\right)$ contributes to finding a trajectory that does at least as well as the expert. $\Phi(\boldsymbol{\psi})$ represents the value of the planned trajectory $\boldsymbol{\psi}$, and $\Phi(\boldsymbol{\psi}^E )$ represents the value of the expert trajectory $\boldsymbol{\boldsymbol{\psi}}^E$.

The trajectory's feature expectation is mapped from the ego vehicle's state and its relative state in the surrounding traffic. The mapping is formulated as follows.

\begin{equation}
    \setlength{\arraycolsep}{3.6pt}
    \begin{array}{c}
    \boldsymbol{\Phi}_{i}(\boldsymbol{\psi})=\boldsymbol{\Phi}_{i}\left(\left[\begin{array}{lllllll}
    s & l & \varphi & \delta & \text{dist}^{p} & \text{dist}^{a}
    \end{array}\right]^{T}\right) \\
    =\left[\begin{array}{llllllllll}
    l_{i} l_{i} & l_{i} & \varphi_{i} \varphi_{i} & \varphi_{i} & l_{i} \delta_{i} & \varphi_{i} \delta_{i} & s_{i} \delta_{i} & \delta_{i} \delta_{i} & \text{dist}_{i}^{p} & \text{dist}_{i}^{a}
    \end{array}\right]^{T} \\
    i \in[1, K] \\
    \end{array}
    \label{Equation34}
\end{equation}
where $\boldsymbol{\Phi}_{i}(\boldsymbol{\psi})$ is the $i^{t h}$ column of $\Phi(\boldsymbol{\psi})$, $s_i$, $l_i$, $\varphi_i$, $\delta_{i}$, $\operatorname{dist}_{i}^{\mathrm{p}}$ and $\operatorname{dist}_{i}^{\mathrm{a}}$ has the same explanations as those variables in equation (\ref{Equation9}) (\ref{Equation10}) (\ref{Equation15}).

Equation (\ref{Equation33}) can be approximated as follows:
\begin{equation}
    \boldsymbol{w}^{(t)}=\min _{\boldsymbol{w}}\left(\boldsymbol{w} \cdot \boldsymbol{\Phi}\left(\boldsymbol{\psi}^{(t-1)}\right)-\boldsymbol{w} \cdot \boldsymbol{\Phi}\left(\boldsymbol{\psi}^{E^{(t)}}\right)\right)
    \label{Equation35}
\end{equation}
\begin{equation}
    \boldsymbol{\psi}^{(t)}=\max _{\boldsymbol{\psi}}\left(\boldsymbol{w}^{(t)} \cdot \Phi(\boldsymbol{\psi})\right)
    \label{Equation36}
\end{equation}
where $\boldsymbol{\psi}^{(t-1)}$ is the trajectory operated in the last iteration. $\boldsymbol{\psi}^{E^{(t)}}$ is the expert trajectory generated in the current iteration. $\boldsymbol{w}^{(t)}$ is the reward matrix updated in the current iteration. $\boldsymbol{\psi}^{(t)}$ is the trajectory generated in the current iteration. Reward correction is realized by solving equation (\ref{Equation35}). Trajectory updating is realized by solving equation (\ref{Equation36}). The detailed process of solving equation (\ref{Equation36}) has been shown in Section \uppercase\expandafter{\romannumeral2}.C.2.

\section{EVALUATION}

The evaluation of the proposed PALC system focuses on two aspects: the effectiveness of personalization and the efficiency of evolution. 

\subsection{Experiment Design}

\subsubsection{Testbed}

A MATLAB-based simulation platform is adopted. A road section with two lanes is set up in the platform. The road section is two-hundred-meter long and the lane is 3.5-meter-width. Background vehicles are created with diverse initial positions and speeds. They follow Intelligent Driver Model (IDM) \cite{treiber2000congested}.

\subsubsection{Test scenario}

Cruising in the traffic, the ego vehicle autonomously makes a lane-changing maneuver when it is impeded. The user would take over when the vehicle's performance deviates from the user's expectations. The threshold for the deviation is 0.3 meters \cite{ziegler2022modeling}.

\subsubsection{Sensitivity analysis}

Sensitivity analysis considers three factors: vehicle's speed, traffic congestion level and the user's driving style.

Four speed types are adopted:

\begin{itemize}
\item {Relative high speed at collector road}: Ego vehicle is with a speed of 45 mph, and the surrounding vehicles are with the speed of 40 mph.
\item {Relative slow speed at collector road}: Ego vehicle is with a speed of 45 mph, and the surrounding vehicles are with the speed of 35 mph.
\item {Relative high speed at major arterial}: Ego vehicle is with a speed of 65 mph, and the surrounding vehicles are with the speed of 60 mph.
\item {Relative slow speed at major arterial}: Ego vehicle is with a speed of 65 mph, and the surrounding vehicles are with the speed of 55 mph.
\end{itemize}

It should be noted that, setting the speeds of the surrounding vehicles lower than that of the ego vehicle is designed to establish a discretionary lane-changing scenario for the ego vehicle. This allows the evaluation to focus on personalization for lane change.

Five types of traffic congestion levels are adopted:

\begin{itemize}
\item Adjacent vehicle headway equals to 50 / 45 / 40 / 35 / 30 meter.
\end{itemize}

Three types of driving styles are adopted \cite{fairclough1997effect} \cite{li2021combined}:

\begin{itemize}
\item {Aggressive}: The user expects a time headway of 1.15 s and a lane change duration of 1.7 s.
\item {Neutral}: The user expects a time headway of 1.23 s and a lane change duration of 2.1 s.
\item {Cautious}: The user expects a time headway of 1.76 s and a lane change duration of 2.5 s.
\end{itemize}

\begin{figure}[thpb]
  \centering
  \includegraphics[scale=0.08]{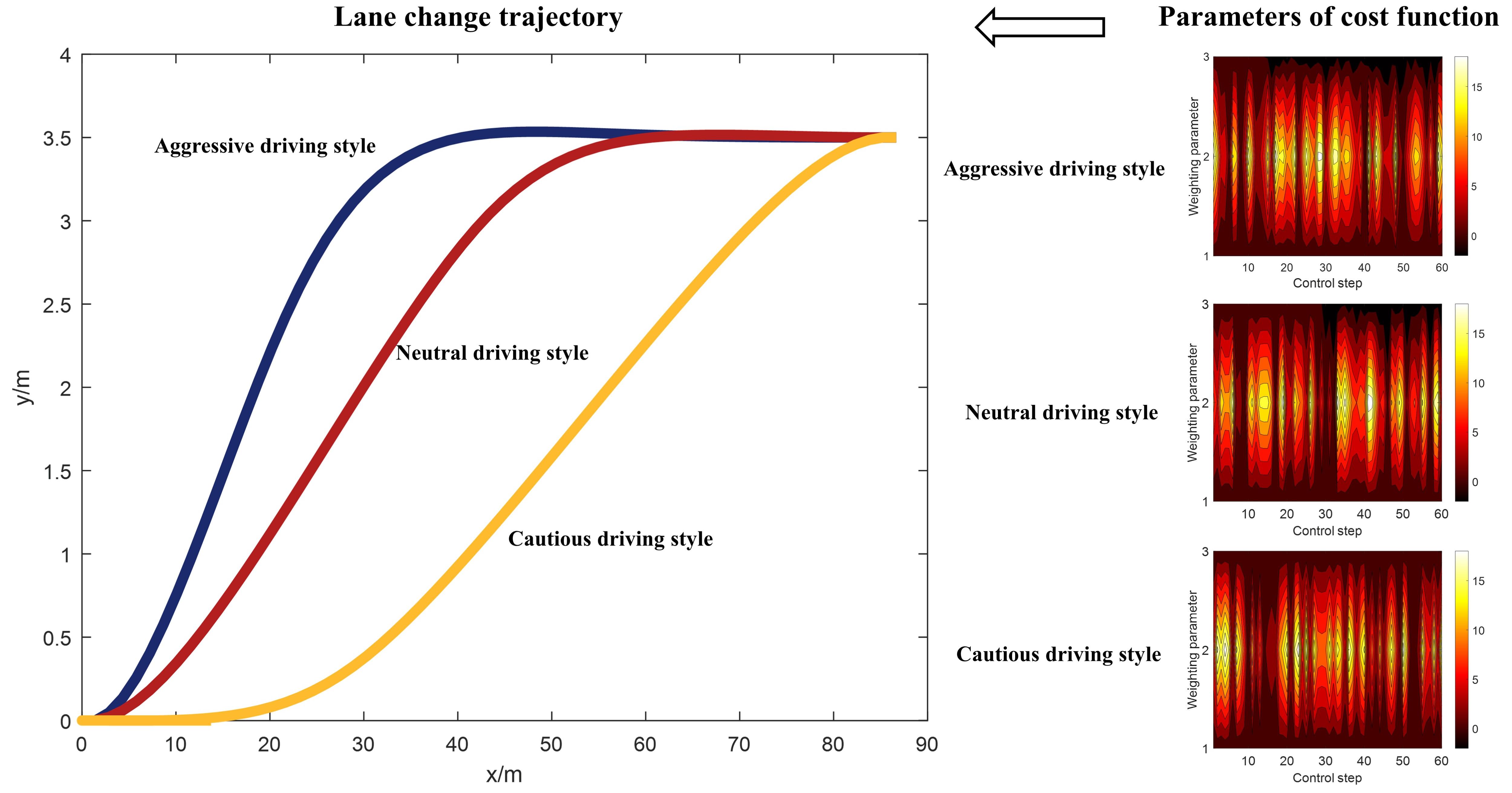}
  \caption{Comparison among different driving styles.}
  \label{Figure5}
\end{figure}

These parameters are validated by implementing the lane-change models in \cite{li2021combined} within the proposed system. The parameters in the cost function of the trajectory planner facilitate lane-changing with different driving styles. The comparison of trajectories and cost function parameters for different driving styles is presented in Fig. \ref{Figure5}.

\subsubsection{Measurements of effectiveness (MOE)}

The proposed PALC system is quantitatively evaluated from evolution efficiency, perceived safety, and computational efficiency. 

\begin{itemize}
\item Evolution efficiency is quantified by the required number of lane changes for customization ($N_{l c}$). 
\item Perceived safety is quantified by the perceived safety distance ratio. Perceived safety distance ratio is defined as the ratio of the distance covered before the takeover to the whole lane-change duration, since intervention has been used as an indicator of perceived safety \cite{tenhundfeld2020trust}.
\item Computational efficiency is quantified by the computation time.
\end{itemize}

\subsection{Results}

The experiment results confirm that the proposed PALC system can achieve the function of personalization, maintaining high evolution efficiency and ensuring perceived safety. The average number of iterations is 13.8, which is 14 times faster than a conventional system. After evolutions, the proposed system is capable of ensuring perceived safety without further takeover interventions. The average computation time is 0.08 seconds, enabling online implementation of the proposed PALC system.

\subsubsection{Function validation}

\begin{figure}[thpb]
  \centering
  \includegraphics[scale=0.16]{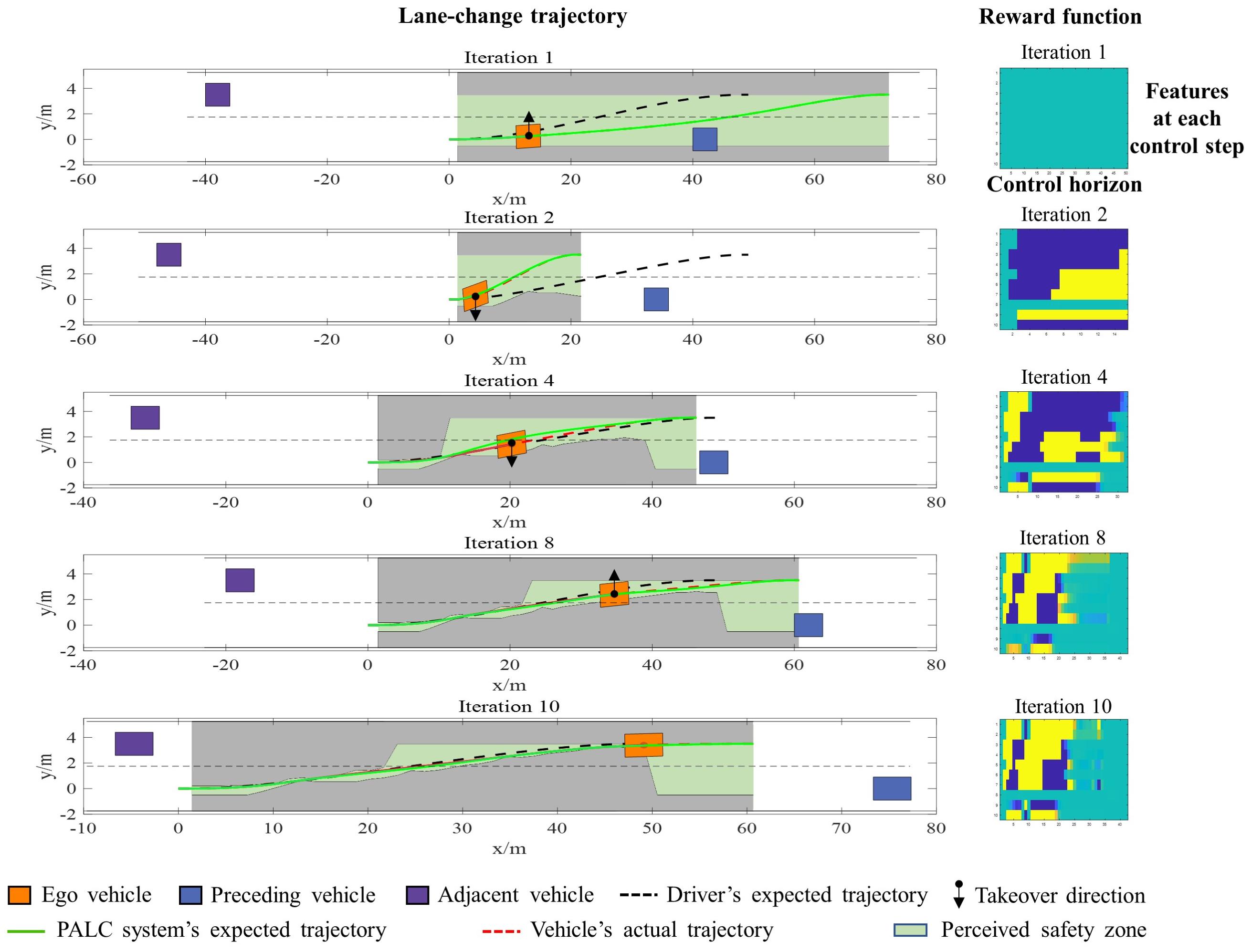}
  \caption{Qualitative example: learning from interventions from an aggressive driver.}
  \label{Figure6}
\end{figure}

The evolution function of the proposed PALC system is verified by the trajectories in Fig.\ref{Figure6}. The vehicles' positions shown in the figure represent their current locations, and the vehicles' trajectories shown in the figure indicate their future trajectories. The visualized matrix represents the weighting matrix of the reward function, where the x-axis corresponds to the control horizon and the y-axis corresponds to the features at each control step. In this example, the driver is more aggressive than the standardized ALC system. ALC system's conservative driving style cannot meet the user's preference. Hence, takeover interventions are conducted by the driver. Learning from the takeover interventions, the PALC system iteratively updates the perceived safety zone and the reward function of the controller, as shown in Fig.\ref{Figure6}. After 10 times of iteration, the proposed PALC system is capable of planning trajectories aligning with the user's expected trajectory, thereby completing the customization. The label “Driver's expected trajectory” in Fig. \ref{Figure6} represents the same concept as “Expected trajectory” in Fig. \ref{Figure2}.

\subsubsection{Evolution efficiency quantification}

\begin{figure}[thpb]
  \centering
  \includegraphics[scale=0.1]{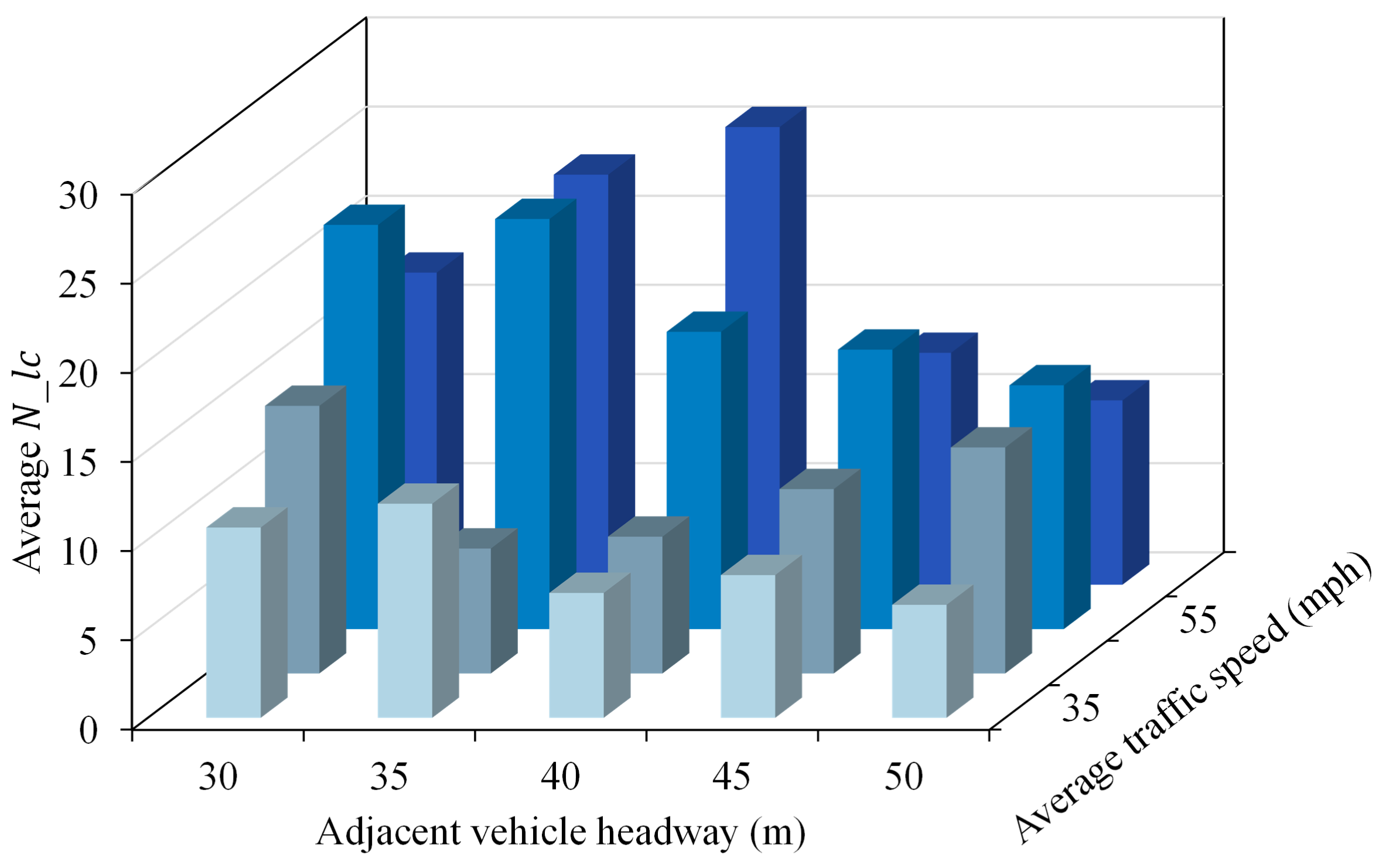}
  \caption{Required number of lane-change before successful personalization.}
  \label{Figure7}
\end{figure}

The number of iterations before a successful personalization is shown in Fig.\ref{Figure7}. A successful personalization means that the ego vehicle completes three consecutive lane changes under identical driving scenarios without requiring human driver intervention. It demonstrates that a user could obtain his customized PALC system with only about 13.8 times of takeover interventions. The evolution efficiency of the proposed PALC method is 13 times faster than traditional PALC systems \cite{yang2021personalized}. The baseline PALC system utilized historical naturalistic driving data to calibrate the parameters of the lane change model. It needs 200 times of lane-changing maneuvers to serve as experimental data. Furthermore, the evolution efficiency is consistent and reliable in all cases, as shown in Fig.\ref{Figure7}. The number of iterations ranges from 6 to 26.

\begin{figure}[thpb]
  \centering
  \includegraphics[scale=0.12]{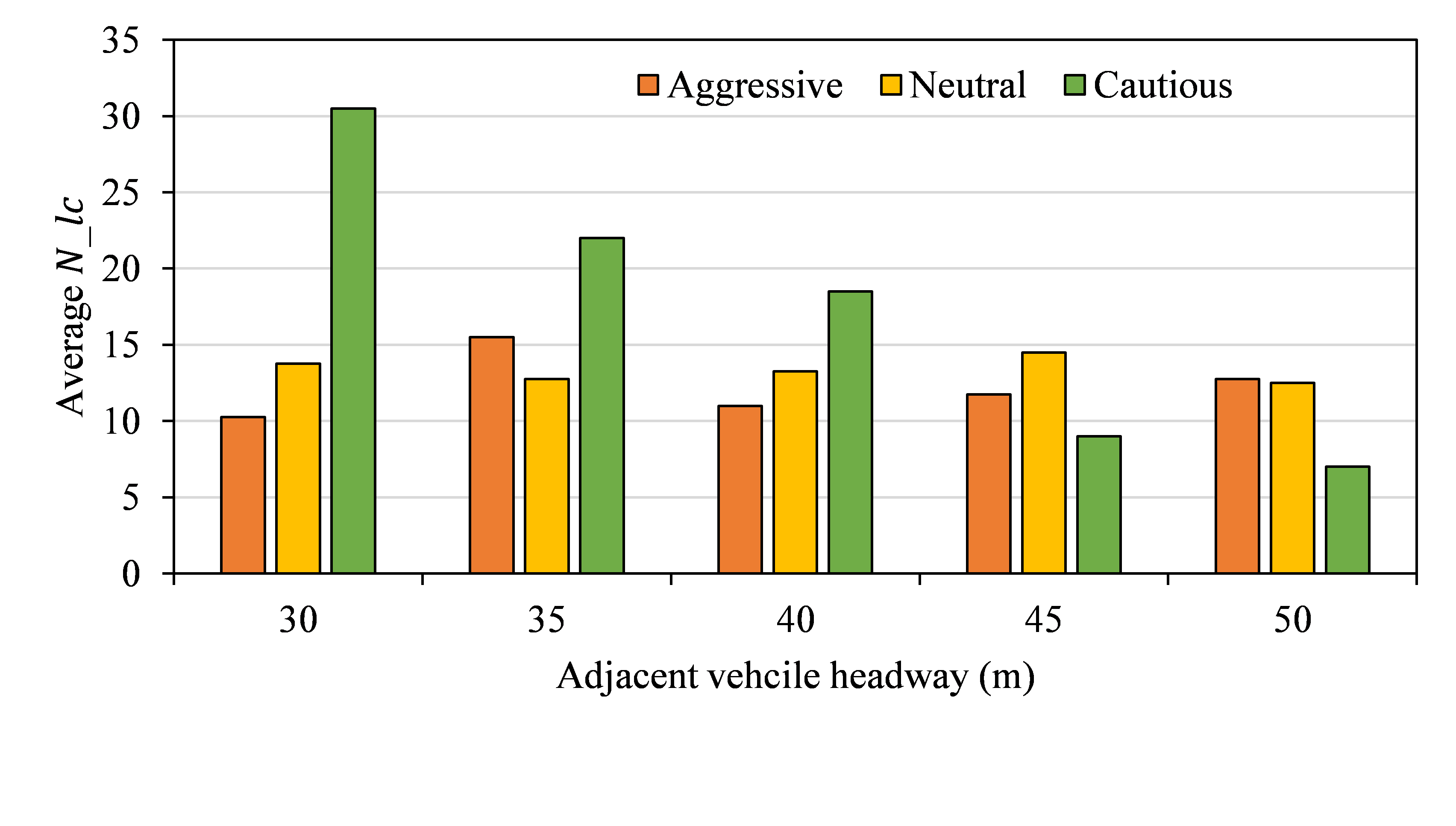}
  \caption{Required number of lane-change with different driving styles.}
  \label{Figure8}
\end{figure}

Another sensitivity analysis is conducted to assess evolution efficiency concerning various driving styles, as shown in Fig.\ref{Figure8}. Results demonstrate that the proposed system has higher evolution efficiency when applied by more aggressive drivers. This phenomenon could be attributed to the tendency of aggressive drivers to intervene more frequently, since they have a lower tolerance for conservative driving behaviors. Consequently, the driving zone can be rapidly narrowed to align with the user's desired parameters. This finding is consistent with the rationale underlying the proposed lesson-learning strategy, which aims to expedite evolution through increased instances of takeovers.

\begin{figure}[thpb]
  \centering
  \includegraphics[scale=0.11]{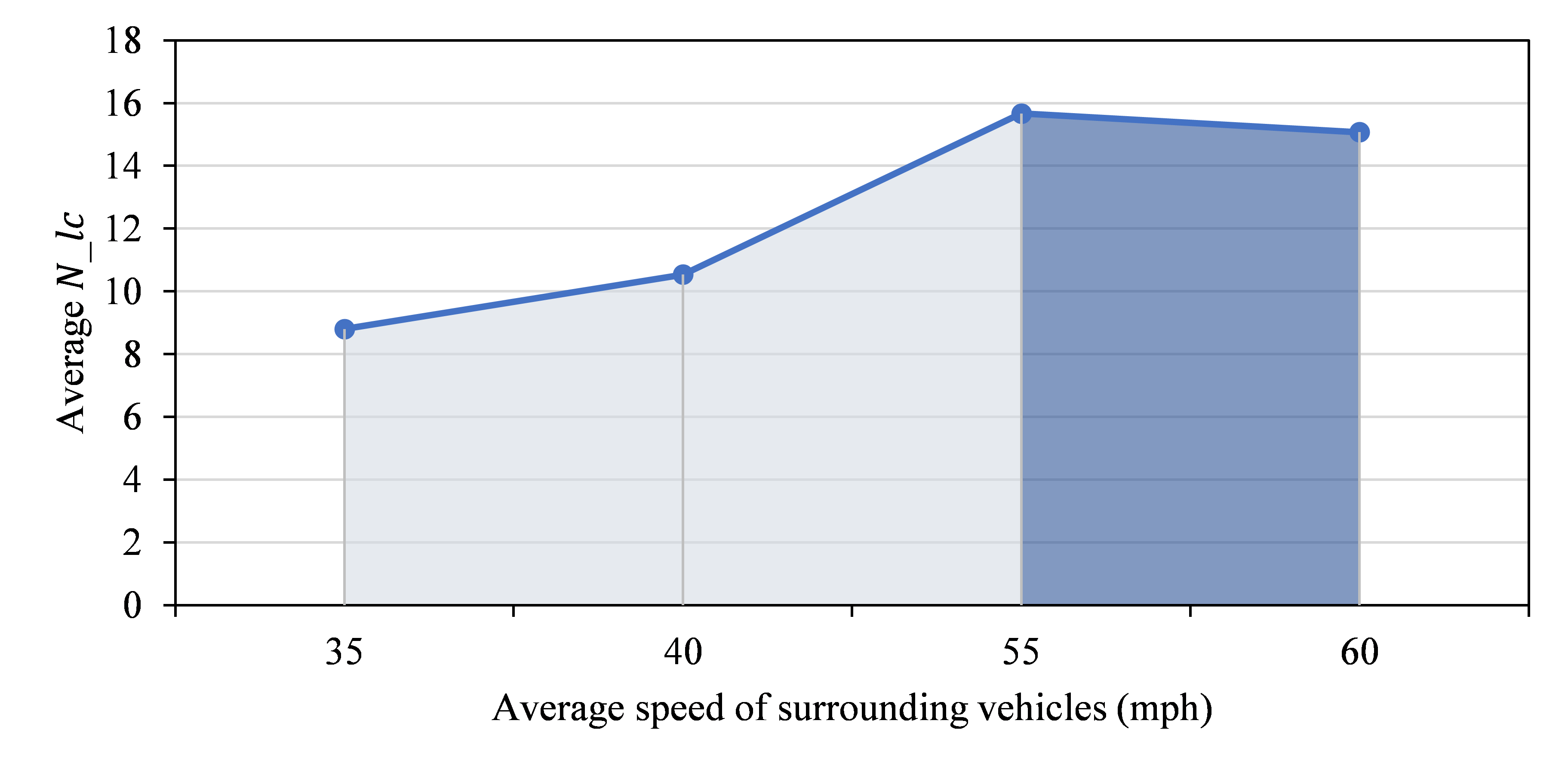}
  \caption{Required number of lane-change with different speeds.}
  \label{Figure9}
\end{figure}

A sensitivity analysis is also conducted to assess the evolution efficiency of the PALC system concerning different traffic speeds, as shown in Fig.\ref{Figure9}. It is demonstrated that the proposed system has higher evolution efficiency when background traffic is slower. It does make sense given that slow-moving surrounding vehicles are less likely to engage in abrupt interactions with the ego vehicle. Only a few iterations are required for the proposed system to align with these steady behaviors. Furthermore, when background traffic becomes faster than 55 mph, the evolution efficiency does not significantly deteriorate. This is because the user tends to take over more cautiously when background traffic speed exceeds 55 mph.

\begin{figure}[thpb]
  \centering
  \includegraphics[scale=0.11]{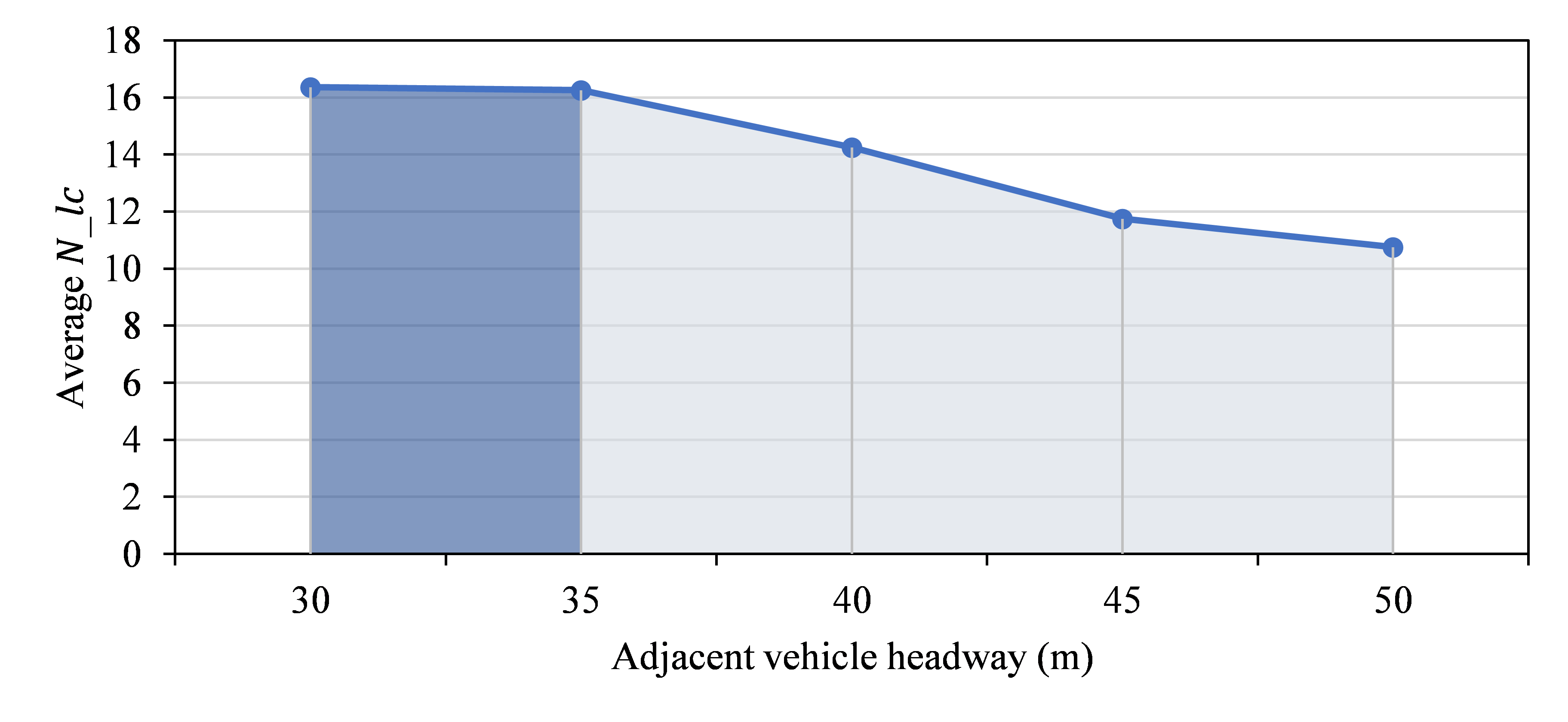}
  \caption{Required number of lane-change with different congestion levels.}
  \label{Figure10}
\end{figure}

In terms of traffic congestion, sensitivity analysis is conducted for evolution efficiency as shown in Fig.\ref{Figure10}. It is demonstrated that the proposed system's evolution efficiency improves with the decrease of congestion level. It makes sense that narrowed spaces in congested traffic may lead to conservative driving style, which significantly deviate from expected trajectories, thereby leading to more times of iterations before customization.

\subsubsection{Experience accumulation function verification}

\begin{figure}[thpb]
  \centering
  \includegraphics[scale=0.1]{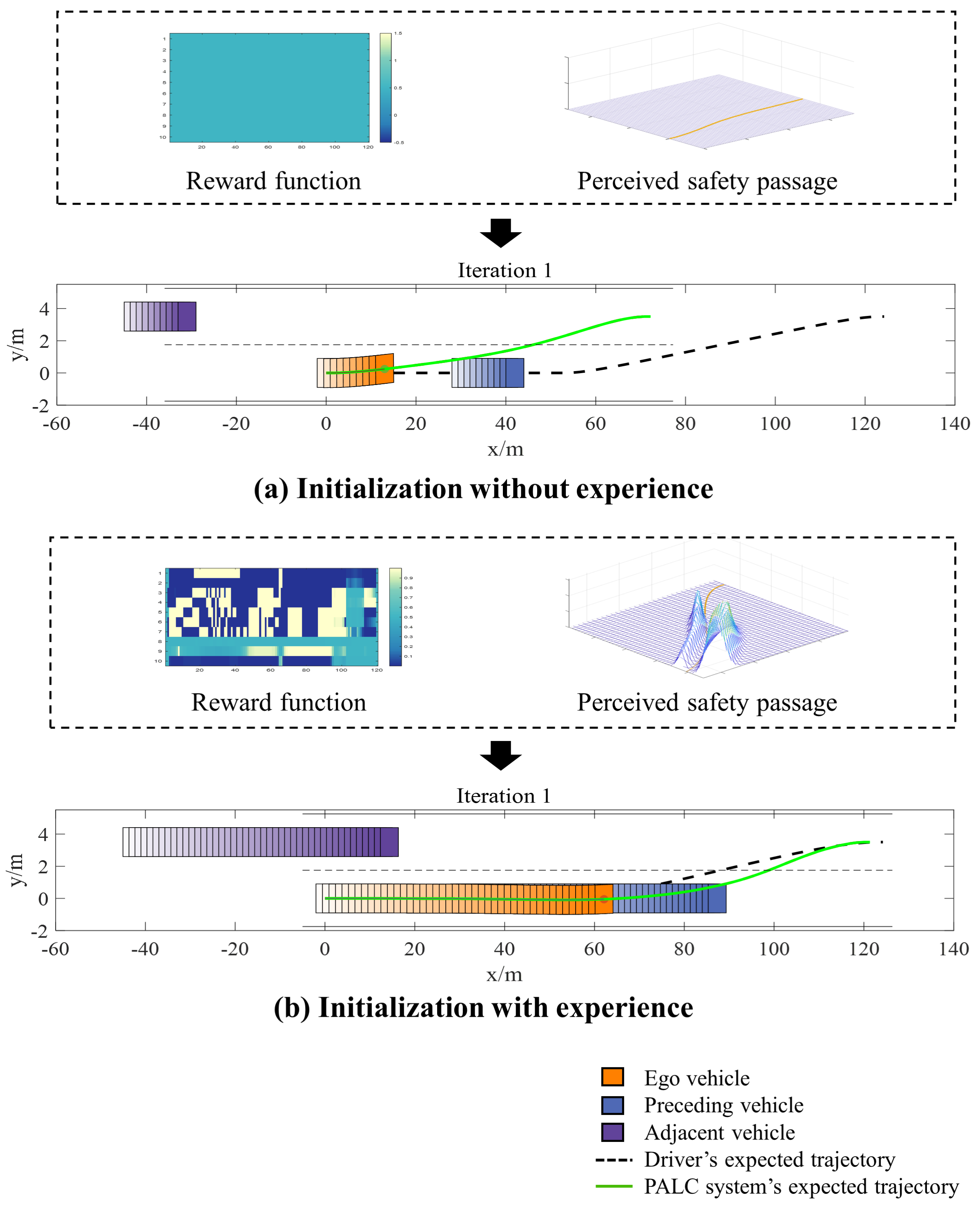}
  \caption{Comparison between with/without accumulated experience.}
  \label{Figure11}
\end{figure}

The experience accumulation capability of the proposed system has been verified. As illustrated in Fig.\ref{Figure11}, when adopted for a new case, the performance of the proposed PALC system is compared between initializing with experience and without experience. It shows that with previous experience, the planned trajectory better aligns with the expected trajectory. However, without experience accumulation, the planned trajectory has a greater bias against the expected trajectory. The experience accumulation capability generally enhances evolution efficiency by 24\% via reducing iterations.

\subsubsection{Perceived safety verification}

\begin{figure}[thpb]
  \centering
  \includegraphics[scale=0.09]{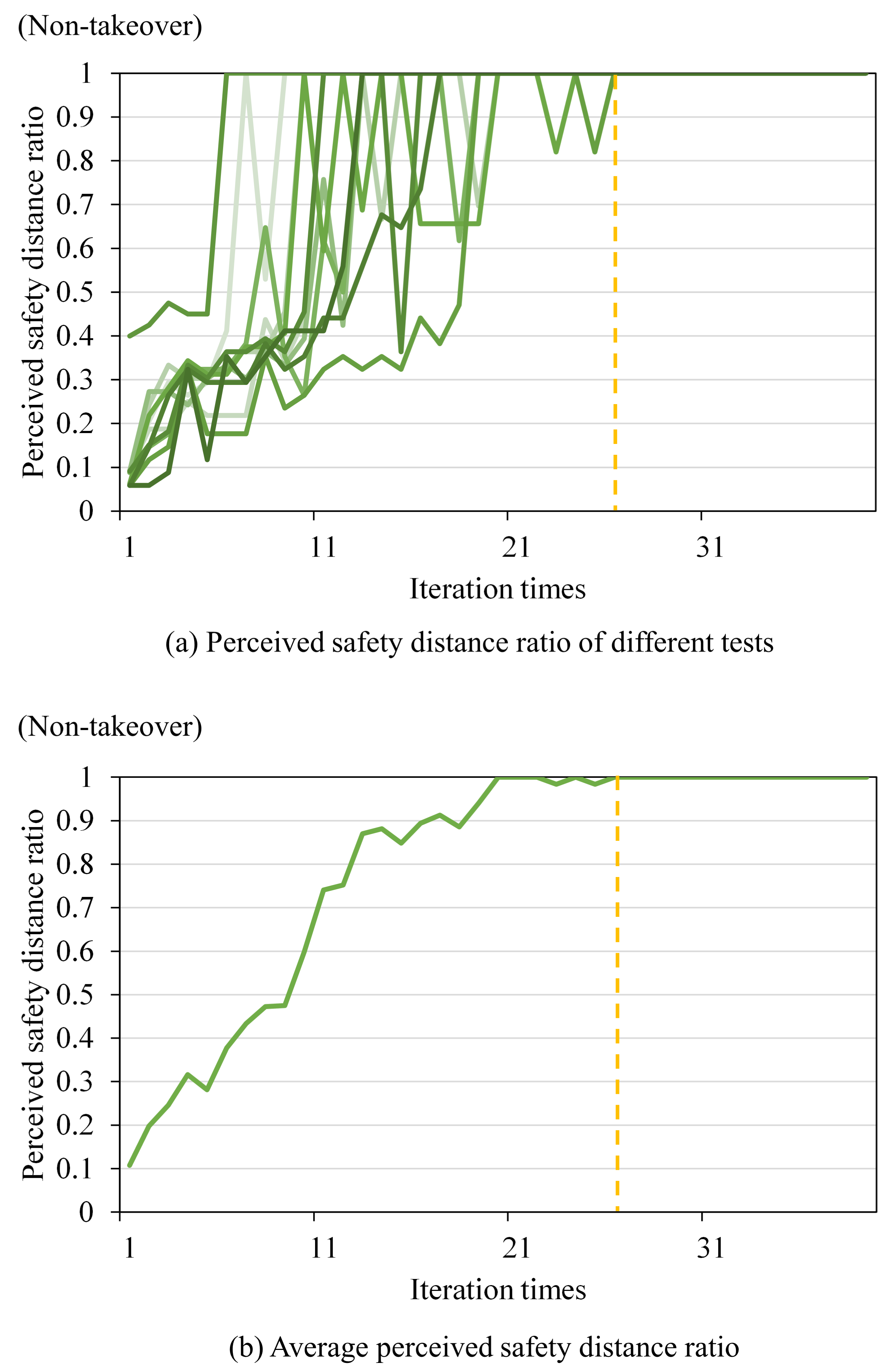}
  \caption{The perceived safety distance ratio changing along with iteration times.}
  \label{Figure12}
\end{figure}

The perceived safety distance ratio is presented in Fig.\ref{Figure12}. It demonstrates that after the required number of iterations, the proposed system is capable of ensuring perceived safety without more takeovers, as shown in Fig.\ref{Figure12}(a). To reach the perceived safe status, the maximum number of required iterations is 26, as shown in Fig.\ref{Figure12}(b). The requirement for iterations is not expected to negatively impact users' adoption of the PADAS system, as it is already more manageable compared to the continuous takeover demands of current commercial ADAS systems. Moreover, these iterations are required only when the model is applied from an original status. When the proposed model is initialized with experience, it could achieve a much faster convergence, as demonstrated in Fig. \ref{Figure11}. This means that the proposed method is generalized enough to be effectively applied across other scenarios. Users do not have to intervene at every case, ensuring the practical utility of the proposed method.

\subsubsection{Computation efficiency quantification}

\begin{figure}[thpb]
  \centering
  \includegraphics[scale=0.09]{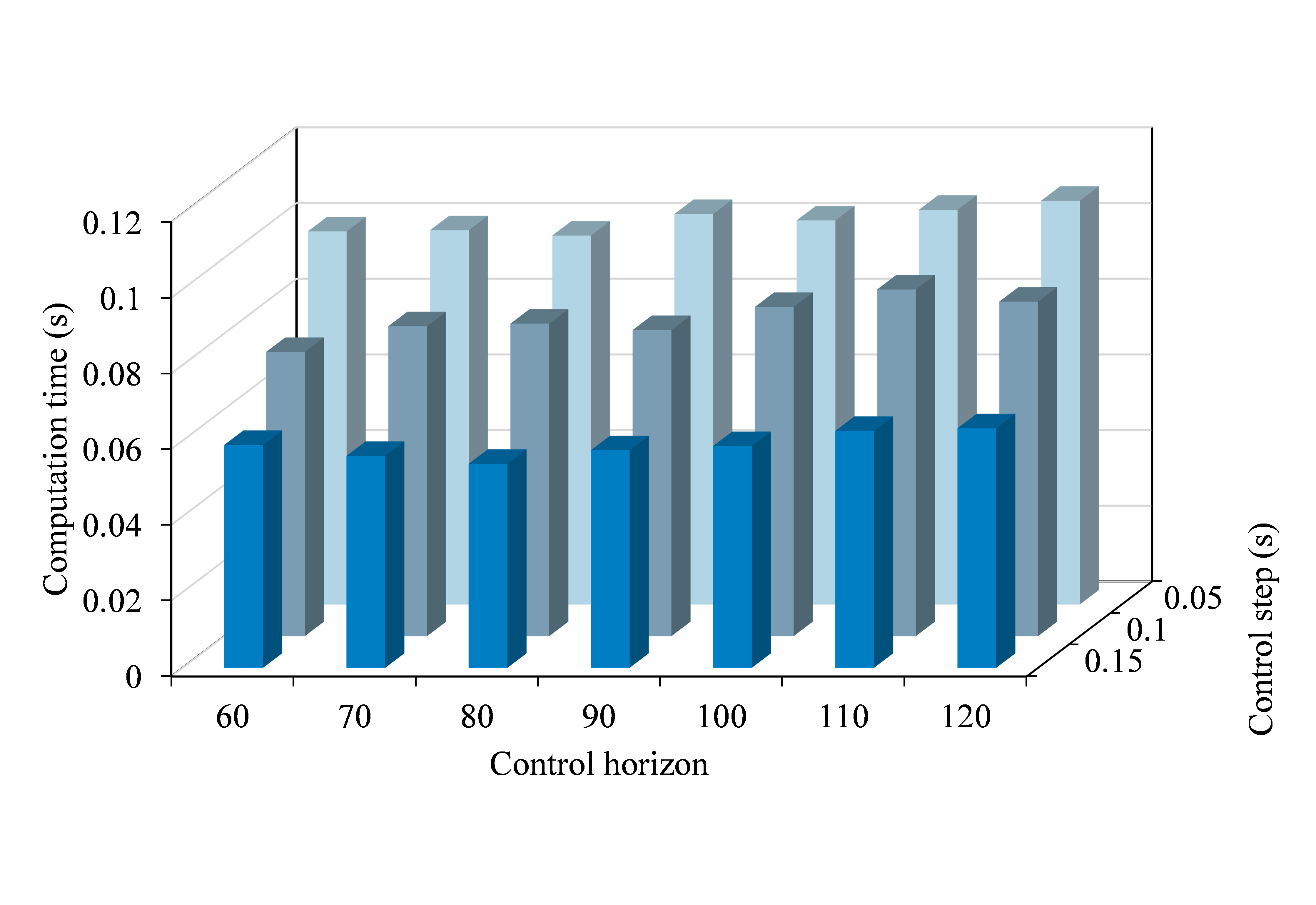}
  \caption{The computation time for an iteration of the proposed PALC system.}
  \label{Figure13}
\end{figure}

The computation time required for an iteration is presented in Fig.\ref{Figure13}. Each iteration takes 0.08 seconds on average. The results demonstrate that increasing the number of parameters, such as using a larger control horizon or a smaller control step, leads to an increase in computation time. However, in all cases, the computation time remains below 0.13 seconds. Furthermore, the most computationally demanding configuration (with a control horizon of 120 steps and a control step of 0.05 seconds) is sufficient to handle lane-change planning effectively. This validates that online implementation of the proposed system can be guaranteed.

\section{CONCLUSION}

This paper proposes a lesson learning based automated lane change controller. It enables online implementation by learning from users' takeover interventions. The proposed method is highlighted for its faster evolution capability, adeptness at experience accumulation, assurance of perceived safety, and computational efficiency. Simulation results demonstrate that:
\begin{itemize}
\item The proposed system consistently achieves successful customization without requiring additional takeover interventions.
\item With an average of only 13.8 learning iterations, the proposed method is 13 times faster than conventional methods \cite{yang2021personalized}.
\item Greater evolution efficiency is observed in more aggressive driving scenarios and in slower, more crowded traffic conditions.
\item Accumulated experience results in a 24\% enhancement in evolution efficiency.
\item The average computation time of 0.08 seconds suggests that the proposed method is well-suited for field implementation.
\end{itemize}

This paper proposes the lesson learning strategy, which may be an enlightenment for the research field of human-like driving. Future studies could access users' historical naturalistic driving data to enrich the training dataset, thereby further reducing the required number of takeovers. As for the generalization objective, future studies could expand the lesson learning concept to more ADAS systems except ALC, such as automated cruise control. Potential improvements could involve transforming the driving zone filter from a longitudinal-lateral space representation to either longitudinal-temporal or longitudinal-lateral-temporal dimensions for better compatibility with control system modeling. Additionally, the proposed system has a potential limitation in handling dense traffic conditions or scenarios involving pedestrians. In such complex environments, drivers must monitor multiple traffic participants under high cognitive load, which could lead to overly conservative takeovers and deviations between the learned perceived safe zones and normal driver behavior ranges. Consequently, the learned control policy may require additional substantial tuning for generalization to other scenarios. To address the limitation, future studies could integrate driver cognition modeling and human-vehicle interaction analysis across diverse traffic environments into the current system. These future research directions contribute to achieving truly universal personalized automated driving.

\section*{ACKNOWLEDGMENT}

This paper is partially supported by National Science and Technology Major 
Project (No. 2022ZD0115504), National Natural Science Foundation of China 
(Grant No. 52302412 and 52372317), Yangtze River Delta Science and 
Technology Innovation Joint Force (No. 2023CSJGG0800), the Fundamental 
Research Funds for the Central Universities, Tongji Zhongte Chair Professor 
Foundation (No. 000000375-2018082), Shanghai Sailing Program (No. 
23YF1449600), Shanghai Post-doctoral Excellence Program (No.2022571), 
China Postdoctoral Science Foundation (No.2022M722405), and the Science 
Fund of State Key Laboratory of Advanced Design and Manufacturing 
Technology for Vehicle (No. 32215011).

\bibliographystyle{IEEEtran}
\bibliography{ref}

%
\begin{IEEEbiography}
[{\includegraphics[width=1in,height=1.25in,clip,keepaspectratio]{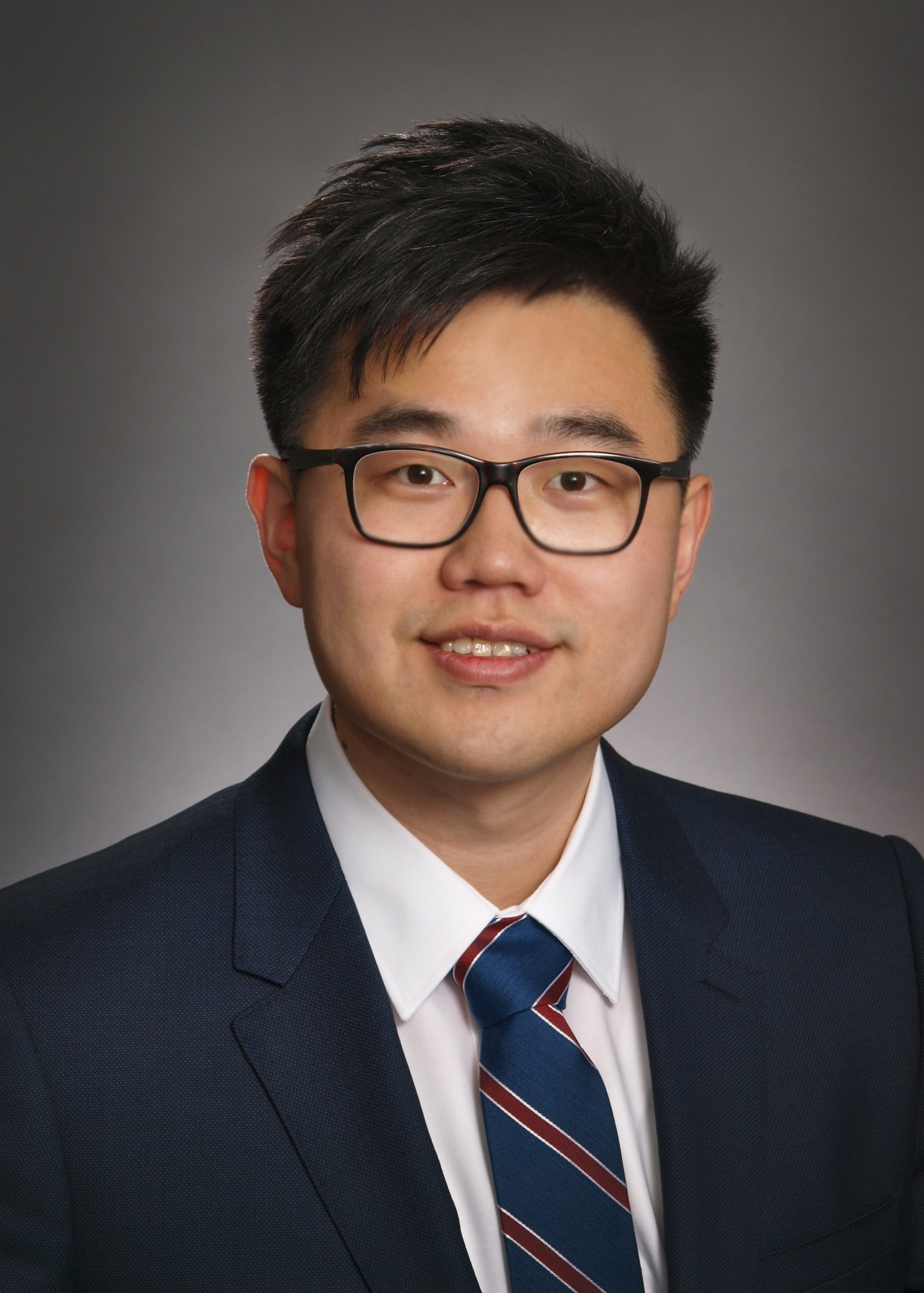}}]{Jia Hu}
(Senior Member, IEEE) is currently working as a Zhongte Distinguished Chair of Cooperative Automation with the College of Transportation Engineering, Tongji University. Before joining Tongji University, he was a Research Associate with the Federal Highway Administration (FHWA), USA. He is an Editorial Board Member of the Journal of Intelligent Transportation Systems and the International Journal of Transportation Science and Technology. He is a member of TRB (a Division of the National Academies) Vehicle Highway Automation Commit-tee, the Freeway Operations Committee, Simulation subcommittee of Traffic Signal Systems Committee, and the Advanced Technologies Committee of the ASCE Transportation and Development Institute. He is the Chair of the Vehicle Automation and Connectivity Committee of the World Transport Convention. He is an Associate Editor of IEEE TRANSACTIONS ON INTELLIGENT VEHICLES, IEEE TRANSACTIONS ON INTELLIGENT TRANSPORTATION SYSTEMS, the American Society of Civil Engineers Journal of Transportation Engineering and IEEE OPEN JOURNAL OF INTELLIGENT TRANSPORTATION SYSTEMS.
\end{IEEEbiography}

\begin{IEEEbiography}
[{\includegraphics[width=1in,height=1.25in,clip,keepaspectratio]{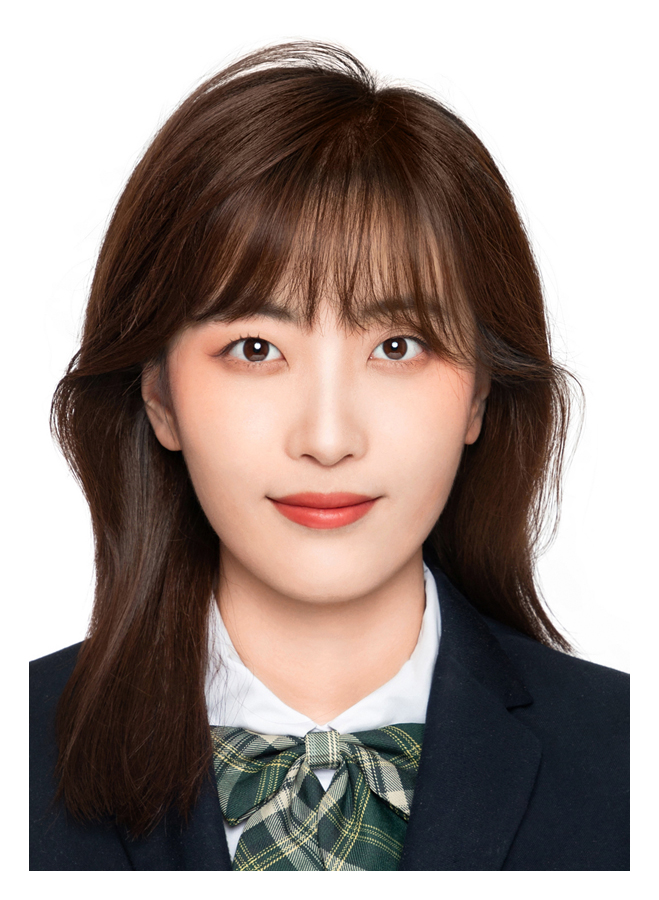}}]{Mingyue Lei}
received the bachelor’s degree in traffic engineering from Southeast University, Nanjing, Jiangsu, China in 2020. She is currently pursuing the ph.D. degree with Tongji University. Her main research interests include optimal control based decision making and motion planning.
\end{IEEEbiography}

\begin{IEEEbiography}
[{\includegraphics[width=1in,height=1.25in,clip,keepaspectratio]{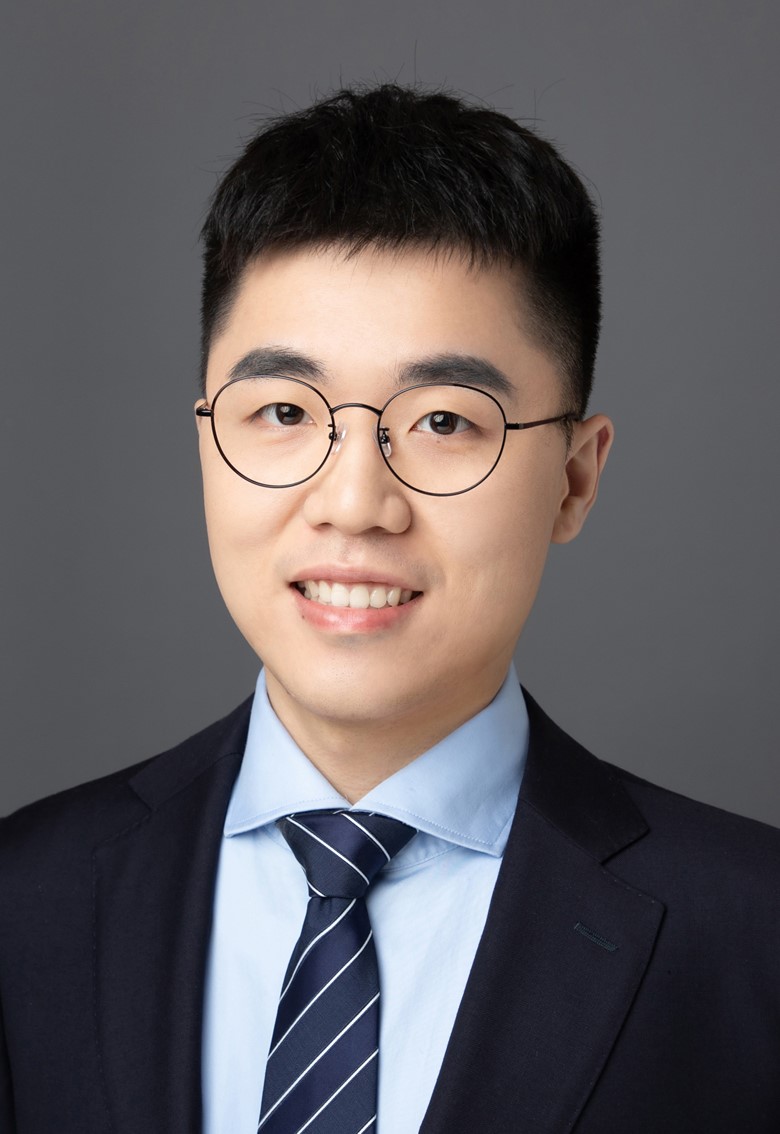}}]{Haoran Wang}
(Member, IEEE) received the bachelor’s degree in transportation engineering from Tongji University, Shanghai, China, in 2017, and the Ph.D. degree from Tongji University in 2022. He is currently a Postdoctoral Researcher with the College of Transportation Engineering, Tongji University. He is a researcher on vehicle engineering, majoring in intelligent vehicle control and cooperative automation.	Dr. Wang served the IEEE TRANSACTIONS ON INTELLIGENT VEHICLES, IEEE TRANSACTIONS ON INTELLIGENT TRANSPORTATION SYSTEMS, Journal of Intelligent Transportation Systems, and IET Intelligent Transport Systems as peer reviewers with a good reputation.
\end{IEEEbiography}

\begin{IEEEbiography}
[{\includegraphics[width=1in,height=1.25in,clip,keepaspectratio]{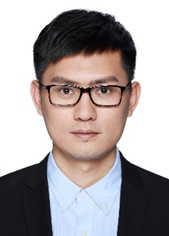}}]{Zeyu Liu}
received the bachelor's degree in Energy and Power Engineering from Beijing Institute of Technology in 2013. He is responsible for the implementation of Baidu Cooperative Vehicle Infrastructure System (CVIS) demonstration projects in Beijing Yizhuang, Wuhan, Shanghai, and other cities. He mainly participated in the compilation of the white paper ``Key Technologies and Prospect of Vehicle Infrastructure Cooperated Autonomous Driving (VICAD) 2.0", and is responsible for the development of the first open-source intelligent connected roadside unit operating system ``Zhilu OS".
\end{IEEEbiography}

\begin{IEEEbiography}
[{\includegraphics[width=1in,height=1.25in,clip,keepaspectratio]{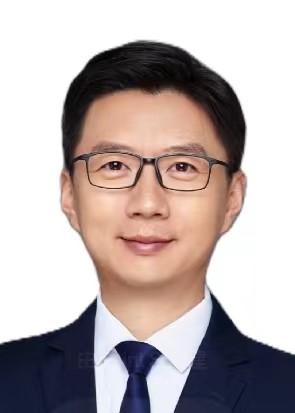}}]{Fan Yang}
received the bachelor’s degree in CS from Beijing University Of Technology , Beijing China in 1999, and the master degree in ICT MOT from Waseda University, Tokyo Japan in 2010. He is currently Chief Architect of Baidu V2X AI Road Platform, Baidu Inc. He is a researcher on vehicle and infrastructure engineering, majoring in intelligent vehicle infrastructure and cooperative automation.
\end{IEEEbiography}







\end{document}